\documentclass[lettersize,journal]{IEEEtran}
\usepackage{amsmath,amsfonts}
\usepackage{amssymb}
\usepackage{booktabs}
\usepackage{algorithmic}
\usepackage{algorithm}
\usepackage{array}
\usepackage[caption=false,font=normalsize,labelfont=sf,textfont=sf]{subfig}
\usepackage{textcomp}
\usepackage{stfloats}
\usepackage{url}
\usepackage{verbatim}
\usepackage{graphicx}
\usepackage{cite}
\begin{document}

\title{Hierarchical Graph Pooling is an Effective Citywide Traffic Condition Prediction Model}

\author{Shilin Pu, Liang Chu, Zhuoran Hou, Jincheng Hu,~\IEEEmembership{Student Member,~IEEE}, Yanjun Huang,~\IEEEmembership{Member,~IEEE}, Yuanjian Zhang,~\IEEEmembership{Member,~IEEE}
        % <-this % stops a space
\thanks{ \emph{(Corresponding author: Yuanjian Zhang.)}

Shilin Pu, Liang Chu, and Zhuoran Hou are with the School of College of Automotive Engineering, Jilin University, Changchun 130022, China (e-mail: pusl20@mails.jlu.edu.cn; chuliang@jlu.edu.cn; houzr20@mails.jlu.edu.cn)

Jincheng Hu is with Department of Aeronautical and Automotive Engineering, Loughborough University, Loughborough, U.K (e-mail: jincheng.hu2020@outlook.com)

Yanjun Huang is with the School of Automotive Studies, Tongji University, Shanghai 201804, China (e-mail: huangyanjun404@gmail.com)

Yuanjian Zhang is with Department of Aeronautical and Automotive Engineering, Loughborough University, Loughborough, U.K (e-mail: y.y.zhang@lboro.ac.uk)
}}

% The paper headers
\markboth{Journal of \LaTeX\ Class Files,~Vol.~14, No.~8, August~2021}%
{Shell \MakeLowercase{\textit{et al.}}: A Sample Article Using IEEEtran.cls for IEEE Journals}

\IEEEpubid{0000--0000/00\$00.00~\copyright~2021 IEEE}
% Remember, if you use this you must call \IEEEpubidadjcol in the second
% column for its text to clear the IEEEpubid mark.

\maketitle

\begin{abstract}
Accurate traffic conditions prediction provides a solid foundation for vehicle-environment coordination and traffic control tasks. Because of the complexity of road network data in spatial distribution and the diversity of deep learning methods, it becomes challenging to effectively define traffic data and adequately capture the complex spatial nonlinear features in the data. This paper applies two hierarchical graph pooling approaches to the traffic prediction task to reduce graph information redundancy. First, this paper verifies the effectiveness of hierarchical graph pooling methods in traffic prediction tasks. The hierarchical graph pooling methods are contrasted with the other baselines on predictive performance. Second, two mainstream hierarchical graph pooling methods, node clustering pooling and node drop pooling, are applied to analyze advantages and weaknesses in traffic prediction. Finally, for the mentioned graph neural networks, this paper compares the predictive effects of different graph network inputs on traffic prediction accuracy. The efficient ways of defining graph networks are analyzed and summarized.
\end{abstract}


\section{Introduction}
\IEEEPARstart{W}{ith} the development of smart cities, the requirements for efficient, continuous, and safe transportation systems are developing strict, resulting in intelligent transportation (ITS) becoming the hotpot \cite{its}. Traffic prediction is an essential component of ITS \cite{zhang1}. Based on the expanding equipment of sensors in intelligent electric vehicles and the development of IoVs facilities, the massive traffic data can be used for predicting general future traffic status to support vehicle-environment coordination control \cite{zhang2}, passenger demand prediction \cite{niu}, trip time estimation \cite{qiu}, and order scheduling \cite{lv}. So aiming at getting accurate traffic status to vehicles, a high-performance traffic predictor is crucial.

Recently, research methods for traffic prediction can be classified into statistical methods \cite{arima, gaussian, bayesian}, basic machine learning methods \cite{svm, ann}, and deep learning methods \cite{yao, zhao, jepsen}. Among them, statistical methods extract the nonlinear features in traffic data by modelling the time series to serve traffic prediction, which includes ARIMA \cite{arima}, Gaussian regression \cite{gaussian}, and Bayesian network \cite{bayesian}. These methods rely on the assumption of time series smoothness. It is challenging to learn the highly nonlinear features in traffic data. Although basic machine learning methods have some nonlinear learning capabilities, such as SVM \cite{svm} and ANN \cite{ann}, they are still not competent for modelling highly spatial and temporal nonlinear traffic data. Their prediction results rely heavily on feature engineering based on expert experience. With the development of deep learning, the method dominates the traffic prediction task under its strong ability to capture temporal- spatial correlations. The mainstream methods are based on convolution neural networks (CNNs) and graph neural networks (GNNs) for spatial correlation capture of traffic data. For extracting temporal correlation, recurrent neural networks (RNNs) and attention mechanisms perform excellently are the mainstream methods.
\IEEEpubidadjcol
Capturing temporal correlation in traffic data focuses on the temporal dimension in the traffic history sequence \cite{xiao}. Besides, the spatial correlation is grabbed by extracting spatial features of traffic data in the spatial distribution dimension. Compared with temporal correlation extraction, there are difficulties with spatial correlation extraction because of the complexity of the definition of spatial data and the large scale of spatial data. Therefore, this paper strengthens the effectiveness of spatial extraction methods, which are used for establishing the model of traffic predictor. The mainstream spatial extraction methods in deep learning are CNNs and GCNs. CNNs, possessing the characteristic of local connectivity and weight sharing, can efficiently grab the spatial correlation in the definition of Euclidean space \cite{toncharoen, yao}. But the regular spatial definition limits the mining of deep-seated information, resulting in the loss of traffic information. In contrast, GCNs, with the input of a graph network, can extract the non-Euclidean correlation, which can construct various graph networks corresponding to the requirement of extraction tasks, to obtain comprehensive spatial correlation \cite{jepsen, ja, tang}. However, GCNs always extracts spatial correlation based on the whole graph network of traffic data. The monotonous scale of the input graph network cannot fully reflect the in-depth information. And the redundant information of the entire graph is gradually inevitable with the increase of the size of the graph. These reasons lead to the deep learning network's inability to capture the spatially features efficiently. Thus, these disadvantages hinder the further improvement of prediction performance. Graph pooling methods \cite{2022Graph} can effectively coarsen graph networks to generate different levels of graph representation. The in-depth graph representation eliminates the redundant information and strengthens the predictive ability of the prediction model by fully exploiting the graph structure information. However, different graph pooling methods have different implementations in terms of ways to achieve graph network coarsening.

\begin{figure}[!t]
\centering
\includegraphics[width=5in]{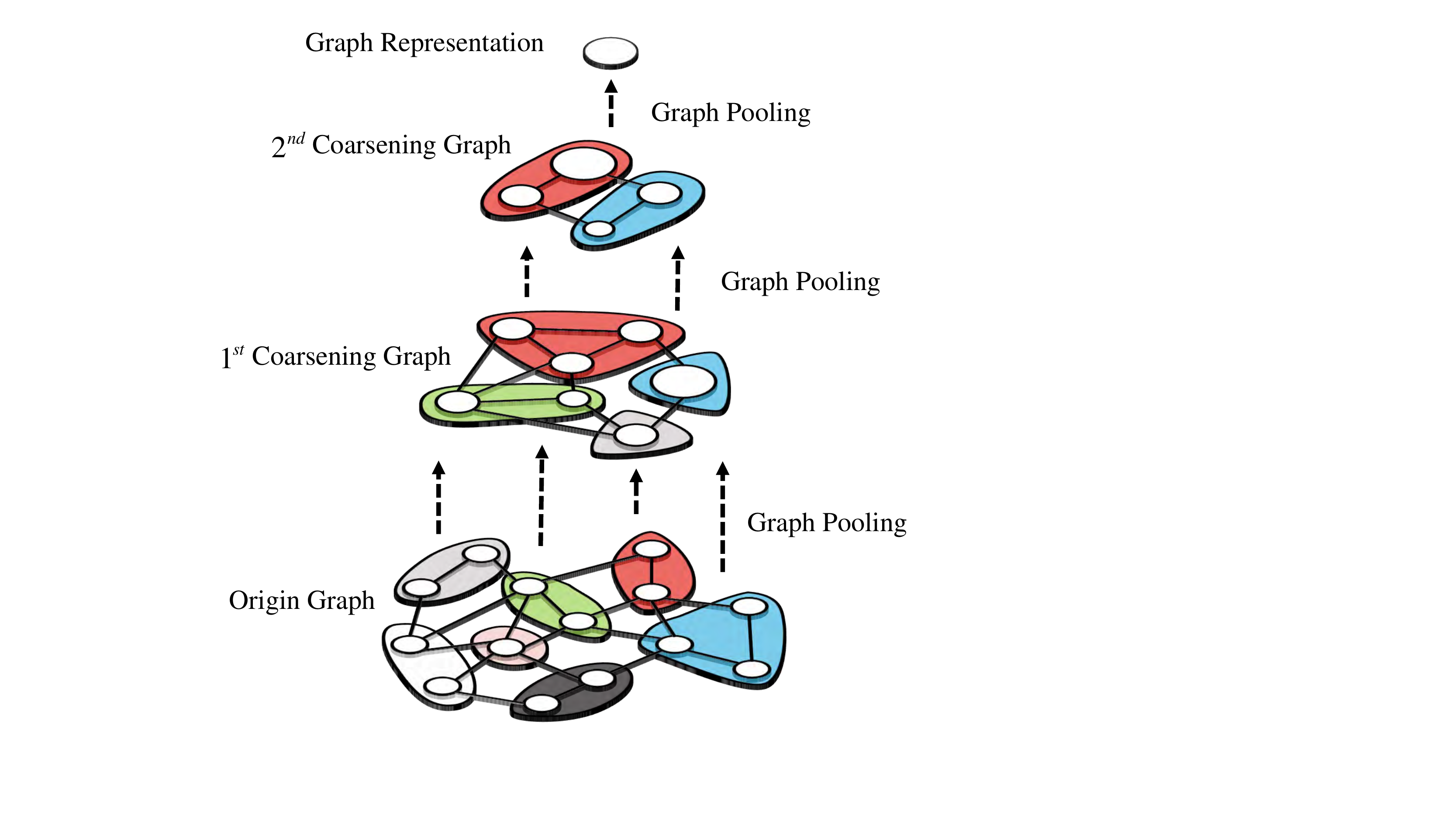}
\caption{Graph network coarsening process for hierarchical graph pooling methods.}
\label{fig_1}
\end{figure}

Graph pooling methods, which are responsible for coarsening the graph network into smaller coarsened graph networks, are mainly classified into planar graph pooling and hierarchical graph pooling according to the different ways of graph-level representation generation. The planar graph pooling method generates the graph-level representation directly by averaging or summing all node embeddings. The Graph network coarsening process of planar graph pooling is over direct by just one step without establishing hierarchical graph representations, resulting in the loss of graph information. The hierarchical graph pooling approach uses node clustering or node dropping to gradually coarsen the graph into smaller ones, as shown in Fig. \ref{fig_1}. The information in different levels of graph representation is learned by retaining valid information. Among them, for the hierarchical pooling method of node clustering, the coarsened graph is constructed by regarding the clusters formed by clustered nodes as new nodes. The classical models include DiffPool \cite{2018Hierarchical} and StructPool \cite{Yuan2020StructPool}. For the hierarchical pooling method of node drop pooling, the coarsened graph is constructed by dropping nodes from the original graph, whose classical models are SAGPool\cite{2019Self} and HGP-SL\cite{2019Hierarchical}. In short, compared with the performance of graph coarsening in planar graph pooling methods, hierarchical graph pooling methods have the ability of multi-level graph representation generation to retain valid traffic spatial information. Thus, this paper focuses on the traffic prediction problem based on hierarchical graph pooling methods. Since hierarchical graph pooling methods generate deep hierarchical graph representation based on different mechanisms, analyzing the performance of different hierarchical graph pooling methods in traffic prediction is necessary.

Different types of hierarchical graph pooling networks reflect various performances in graph representation extraction and coarsening methods. Firstly, comparing the performance different hierarchical graph pooling networks in traffic prediction tasks is necessary. Accordingly, graph pooling methods are compared with mainstream deep learning methods, aiming at analyzing the feature extracting ability of each method. Secondly, the hierarchical graph pooling methods, including node clustering pooling and node drop pooling, are adopted to establish traffic predictor and the performance of the methods are compared to weight the impact on the accuracy of traffic prediction. Finally, various ways for generating graph networks are integrated in the traffic predictor, which directly influence on the effect of traffic predictor. The detailed contribution added to the existing literature include:

\begin{enumerate}
\item{The graph pooling methods are adopted in traffic predictor, which reduces the redundant information of the raw data to strengthen the capacity of extracting spatial correlation. And the validation using real-world transportation datasets demonstrates the state-of-the-art performance in the traffic prediction task.}
\item{Two classical hierarchical graph pooling methods are compared by analyzing the performances of the corresponding traffic predictors, whose evaluation includes the accuracy and quadratic weighted kappa coefficient. The adaptiveness of hierarchical graph pooling methods is explored in traffic prediction.}
\item{Four graph network establishing ways from various perspectives are discussed in this paper, whose constructing concepts include the aspects of topology, geography, historical pattern, and road attributes. The analysis gives guidance for increasing the predictive capacity of traffic predictors.}
\end{enumerate}

The rest of the paper is organized as follows: Section II describes in detail the current status of the application of deep learning methods to traffic forecasting tasks and the development of graph pooling methods. Section III introduces the data definition in traffic prediction tasks and specifies the objectives of traffic level prediction tasks. Section IV describes in detail two classical hierarchical graph pooling methods based on a summarized hierarchical graph pooling architecture. Section V describes in detail the test experiments of graph pooling methods in traffic forecasting tasks. Section VI analyzes the positive effects of hierarchical graph pooling methods with different representations of graph networks for the traffic forecasting task based on the experimental results. Finally, Section VII outlines the conclusions and future perspectives.

\section{Related Work}
Deep learning is regarded as an effective solution in solving the modeling problem of complex nonlinear and spatiotemporal correlation characteristics of traffic data. Various deep learning methods are proposed for capturing temporal and spatial correlations of traffic data. For spatial correlation extraction, GNNs are widely used. However, the commonly used GNN, i.e., Graph Convolution Networks (GCNs), suffers from the problems of the monotonous scale of graph networks and information redundancy in large-scale graph networks, resulting in the limitation of the further improvement of prediction accuracy. Graph pooling networks, as the development trend of GNNs, can extract multi-level graph representations and reduce redundant information through graph coarsening operation. To be specific, various graph pooling methods perform graph coarsening operation under different mechanisms, so it is necessary to discuss the performance of different graph pooling methods in traffic prediction tasks.

Capturing the spatio-temporal correlations of traffic data is the key to traffic predictive tasks. The research on capturing nonlinear temporal correlation of traffic data is mainly based on RNN and its variants, i.e., Long Short-Term Memory (LSTM) or Gated Recurrent Unit (GRU). Xiao et al. captured the forward and reverse trends of traffic state sequences based on multi-resolution (daily and weekly) traffic data by using bi-directional LSTM (Bi-LSTM) \cite{xiao}. Tian et al. used LSTM to capture temporal correlations in short-term prediction tasks of traffic flow \cite{tian}. However, RNNs-like models are more often applied in short-term prediction and suffer from information loss and low computational efficiency when facing the problem of feature extraction for long-time sequences.

For spatial correlation extraction of traffic data, earlier researchers focus more on CNNs, which have the advantages of capturing Euclidean spatial correlation features because of local connectivity and weight sharing. Toncharoen et al. performed spatial correlation extraction from nodes along highways by CNNs \cite{toncharoen}. Yao et al. used CNNs to capture spatial correlation for traffic data distributed in the form of regions \cite{yao}. However, traffic data presents more irregular distribution forms, and GCNs focusing on non-Euclidean spatial correlation extraction are gradually becoming mainstream compared to CNNs. Specifically, GCNs capture the non-Euclidean spatial correlation from graph networks, which are defined by regarding the roads or areas as nodes and considering the connection relationships between roads or areas as edges. Zhao et al. used GCN to learn complex topologies to capture spatial correlation and excel in traffic prediction tasks of urban road networks \cite{zhao}. Jepsen et al. improved GCN to complete the fusion of three road network attributes, i.e., intersection, road segment, and attributes between road segments. Then, the model is applied to travel speed estimation and speed limit classification of roads \cite{jepsen}. Tang et al. and Ja et al. introduced an adaptive adjacency matrix mechanism in GCN to capture the spatial correlation of traffic data. The formulation of the adjacency matrix was based on the dynamic adjustment mechanism of spatial association \cite{ja}\cite{tang}. However, the GCN method, based on a single-scale graph network, can hardly sieve out the redundant information in large-scale graph networks and learn more profound graph-level representations. This phenomenon limits the prediction performance improvement of GCN-like methods. But with the appearance of graph pooling methods, these problems can be alleviated.

Graph pooling methods are classified into planar graph pooling and hierarchical graph pooling according to graph-level representation generating methods. The planar graph pooling generates the graph-level representation, i.e. the feature vector denoting the entire graph network, directly in one step. Besides, the hierarchical graph pooling gradually coarsens the graph network through a multi-step operation. For planar graph pooling methods, the most common methods are summation pooling and averaging pooling, performing summation or averaging operations on overall node representations \cite{2015Convolutional}. Moreover, based on summation pooling and averaging pooling, some researchers perform additional nonlinear transformations to improve the expressiveness of pooling methods \cite{2019Universal, 2021DropGNN}. Some other scholars introduce soft attention mechanisms to determine the critical weight of each node in the graph-level representation \cite{2019Structured, 2022Multi}. However, most planar pooling methods coarsen the original graph network directly into graph-level representation without considering the hierarchical structure of graphs, leading to information loss in the graph network and reduce the representational power of the graph. 

Hierarchical pooling preserves the structural information of graphs at different levels by coarsening the original graph network into smaller ones. The categories of it can be divided into node clustering pooling and node drop pooling according to the coarsening method. For node clustering pooling, similar nodes are mapped into a cluster by a clustering method, and the clustered clusters are considered as the nodes of new graph. Among them, reasonable clustering methods have become the focus of scholars' research. DiffPool\cite{2018Hierarchical} is a classical clustering pooling method, using the GCN to learn the cluster assignment matrix to complete node clustering. StructPool\cite{Yuan2020StructPool} extends the higher-order structural relationship by showing and capturing DiffPool. LaPool\cite{2019Towards} and MinCutPool\cite{Bianchi2020Spectral} design the assignment matrix from a spectral clustering perspective. MemPool\cite{2020Memory} generates the clustering assignment matrix using a clustering-friendly distribution.

Node drop pooling uses a learnable scoring function to remove nodes with low significance scores to form a new coarsened graph from the remaining nodes. For node drop pooling methods, scholars prefer to design a more sophisticated score generator to select representative nodes, thus leading to retain important graph information. topKPool \cite{2019Graph} is the earliest node drop pooling method for obtaining node scores by learning a mapping matrix. SAGPool \cite{2019Self} learn the saliency score of each node based on a graph convolutional network. HGP-SL \cite{2019Hierarchical} obtains the information score by calculating the Manhattan distance between the target and neighbor node representation. GSAPool \cite{2020Structure} and TAPool \cite{2021Topology} generate node scores from local and global perspectives. The selected and non-selected nodes determine the generated graphs to reserve the valid information, while node drop pooling method inevitably has the problem of information loss. However, relying on an efficient graph coarsening approach, the node drop pooling method is suitable for large-scale graph networks.

\section{Preliminary}

\subsection{Definition 1: Traffic Condition Sample}
The traffic condition signals (e.g., average speed, volume, and occupancy) of all roads on the traffic network at time $t$ are represented as a matrix ${{x}_{t}}\in {{\mathbb{R}}^{n\times m}},t\in [1,\cdots ,T]$, where $n$ represents the number of roads, $m$ indicates the dimension of road characteristics, and $T$ denotes the total timestamp length. Then the traffic condition samples are represented as ${{S}_{t}}=({{X}_{t}},{{Y}_{t}})$, where the input ${{X}_{t}}=[{{x}_{t-w+1}},\cdots ,{{x}_{t-1}},{{x}_{t}}]$ denotes the sequence of historical status signals of length $w$ . The output is the traffic condition grade ${{Y}_{t}}=Cl{{s}_{t+h}}\in {{\mathbb{R}}^{n}}$ of all roads at the timestamp $t+h$, to be specific, the traffic class is described in detail in Definition 3. For all traffic data, the traffic condition sample sets are denoted as $S=[{{S}_{1}},\cdots ,{{S}_{N}}]$, where $N$ denotes the total number of sample sets.

\subsection{Definition 2: Road Graph Network}
Traffic networks exhibit non-Euclidean properties in spatial distribution. Graph networks, which can define the non-Euclidean spatial correlation of road network effectively, are utilized widely in transportation [].In this paper, the road graph network is represented as an undirected graph $G=(V,E,W)$, where each node ${{v}_{i}}\in \ |V|,i\in N$ denotes a road and $N$ represents the total number of roads; each edge ${{e}_{ij}}\in E$ denotes the connectivity of a road ${{v}_{i}}$ with a road ${{v}_{j}}$; the weight ${{w}_{ij}}\in W$ of an edge ${{e}_{ij}}$ represents the degree of correlation between road ${{v}_{i}}$ and road  , to be specific, the larger the weight indicates the stronger spatial correlation between two roads.

Since road graph networks have different structures, the presentations of spatial correlation characteristics are subsequently different. GNNs based on different types of graph networks show heterogeneity in extracting the spatial correlation of traffic data. A reasonable graph network structure can adequately encode the spatial correlation among nodes, thus exhibiting outstanding predictive performance. Based on previous studies \cite{noneuropean}, this paper constructs graph networks in four aspects: topological distribution graph ${{G}_{r}}$, geographical distribution graph ${{G}_{g}}$, historical pattern graph ${{G}_{p}}$, and road attribute similarity graph ${{G}_{a}}$.

\subsubsection{Topological Distribution Graph Network}
\begin{figure}[!t]
\centering
\includegraphics[width=3.5in]{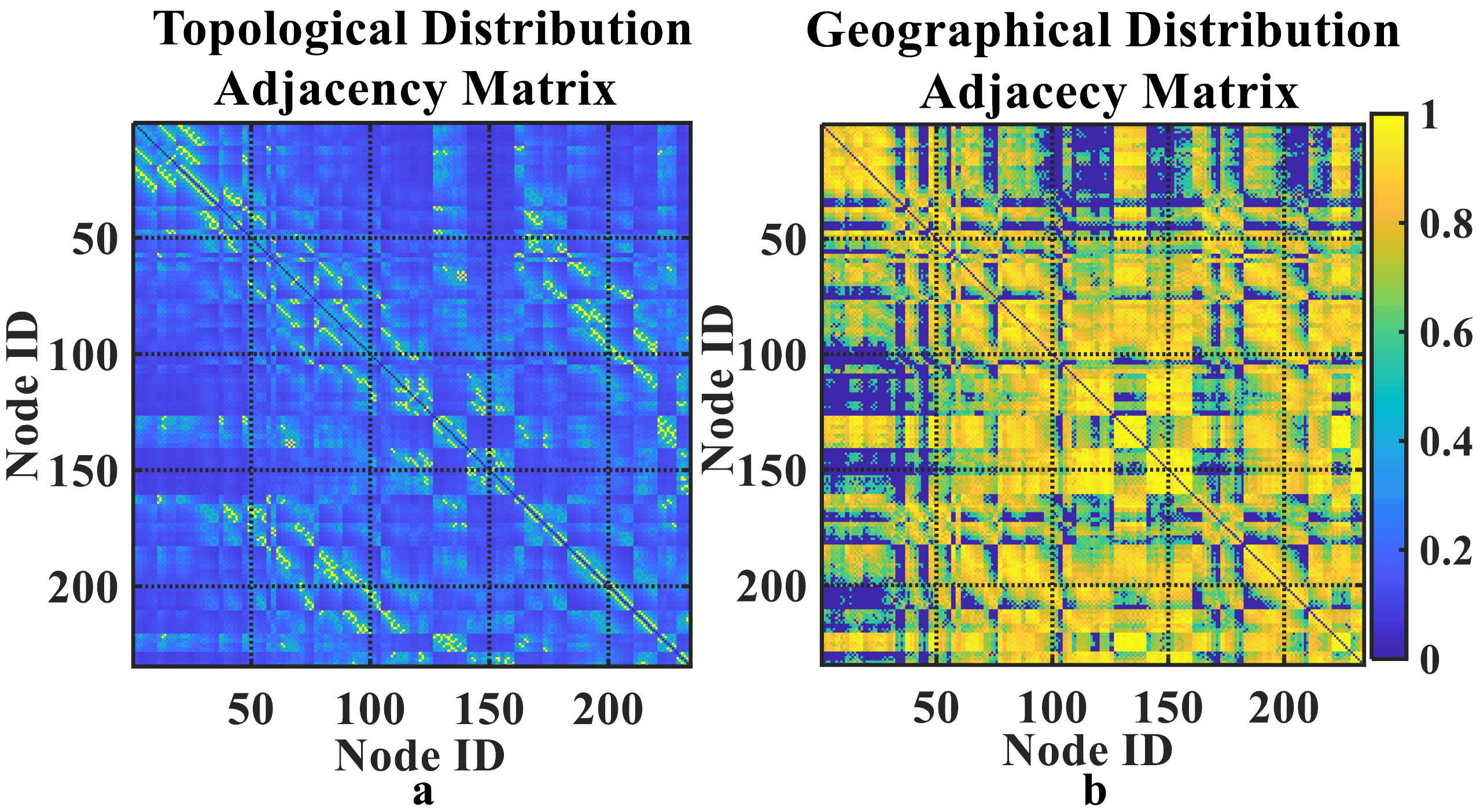}
\caption{Topological Distribution Adjacency Matrix and Geographical Distribution Adjacency Matrix.}
\label{fig_2}
\end{figure}

The topological distribution of a road network is an important characteristic of the road network, which directly reflects the spatial distribution among roads. The road network topology can be expressed as an undirected graph ${{G}_{r}}=(V,E,{{W}_{r}})$, where the weights ${{\omega }_{r}}(i,j)$ of the edges ${{e}_{ij}}$ represent the inverse of the number of edges passed in the shortest path between the road ${{v}_{i}}$ and the road ${{v}_{j}}$. Specifically, the correlation coefficient between roads with fewer edges apart is larger, i.e., the greater the weight in the adjacency matrix. The adjacency matrix ${{W}_{r}}$ of ${{G}_{r}}$ is expressed as Eq. \eqref{eq:1}. The adjacency matrix is visualized as shown in Fig. \ref{fig_2}. a.

\begin{equation}
\label{eq:1}
{{W}_{r}}=\left[ \begin{matrix}
   0 & {{\omega }_{r}}(1,2) & \cdots  & {{\omega }_{r}}(1,N)  \\
   {{\omega }_{r}}(2,1) & 0 & \cdots  & {{\omega }_{r}}(2,N)  \\
   \vdots  & \vdots  & \ddots  & \vdots   \\
   {{\omega }_{r}}(N,1) & {{\omega }_{r}}(N,2) & \cdots  & 0  \\
\end{matrix} \right]
\end{equation}

\subsubsection{Geographical Distribution Graph Network}
In the distribution of real road networks, there is a phenomenon that the two roads with fewer intermediate intervening roads are geographically distant from each other, representing a weak spatial correlation between them. Besides, the phenomenon could cause information misjudgment, which is not conducive to reflecting the actual spatial distribution. To avoid the adverse impact of the phenomenon, which is extracting spatial correlation by simple topological relationship alone, combining the distance among nodes with original topological relations is significant.

The geographical distribution graph is expressed as an undirected graph ${{G}_{g}}=(V,E,{{W}_{g}})$, and the weights ${{\omega }_{g}}(i,j)$ of the edges ${{e}_{ij}}$ are calculated as in Eq. \eqref{eq:2}, where $len({{v}_{m}})$ denotes the sum of the lengths of the roads covered in the shortest path between the road ${{v}_{i}}$ and the road ${{v}_{j}}$ in the road network. The adjacency ${{W}_{g}}$ matrix of ${{G}_{g}}$ is expressed as Eq. \eqref{eq:3} and the matrix is visualized as shown in Fig. \ref{fig_2}. b.

\begin{equation}
\label{eq:2}
{{\omega }_{g}}(i,j)=\frac{len({{v}_{i}})+len({{v}_{j}})}{len({{v}_{m}})}
\end{equation}

\begin{equation}
\label{eq:3}
{{W}_{g}}=\left[ \begin{matrix}
   0 & {{\omega }_{g}}(1,2) & \cdots  & {{\omega }_{g}}(1,N)  \\
   {{\omega }_{g}}(2,1) & 0 & \cdots  & {{\omega }_{g}}(2,N)  \\
   \vdots  & \vdots  & \ddots  & \vdots   \\
   {{\omega }_{g}}(N,1) & {{\omega }_{g}}(N,2) & \cdots  & 0  \\
\end{matrix} \right]
\end{equation}

\subsubsection{Historical Pattern Graph Network}
Compared to topological distribution and geographic distribution in road network, the similarity of road historical patterns, i.e. traffic historical conditions, indirectly affects the spatial correlation. The similarity distribution is reflected in a phenomenon that the two roads in different office areas show similar traffic patterns in the morning and evening peak periods. It is worth exploring to measure spatial distribution from a functional perspective by capturing pattern similarity. The traffic history pattern graph is defined as an undirected graph ${{G}_{p}}=(V,E,{{W}_{p}})$, where the weights ${{\omega }_{p}}(i,j)$ of the edges ${{e}_{ij}}$ are the similarity between the road ${{v}_{i}}$ and the road ${{v}_{j}}$ in terms of historical traffic patterns, and the similarity weights are calculated as follows according to the historical traffic states of different roads.

\begin{figure}[!t]
\centering
\includegraphics[width=3.5in]{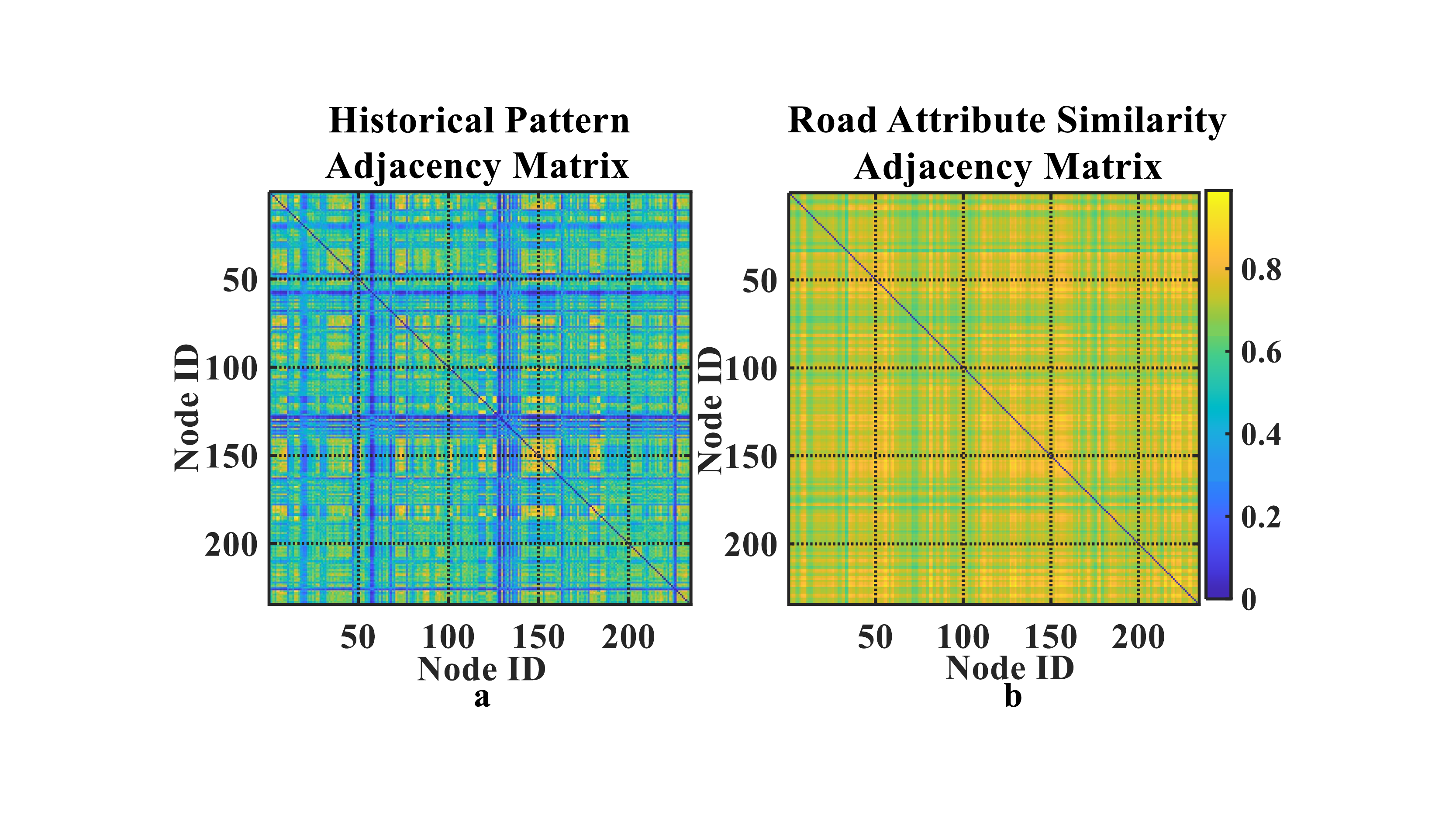}
\caption{Historical Pattern Adjacency Matrix and Attribute Similarity Adjacency Matrix.}
\label{fig_3}
\end{figure}

First, based on the full graph traffic data at hourly resolution (e.g., traffic flow), given the road ${{v}_{i}}$ at the time $t$, a sequence $pv_{t}^{i}\in {{\mathbb{R}}^{\Delta t}}$ of traffic conditions can be developed during the historical time $\Delta t$. Second, given sequences $sv_{t}^{i}$ and $sv_{t}^{j}$ of historical traffic patterns for any two roads ${{v}_{i}}$ and ${{v}_{j}}$, the similarity distance between the two historical patterns is calculated by using DTW (Dynamic Time Warping), denoted as $dis{{t}_{t}}(i,j)$; then, the similarity distance is transformed into the correlation weight by the Eq. \eqref{eq:4}, denoting as ${{\omega }_{p,t}}(i,j)$. In the formula, $\alpha \in [0,1]$ is used to express the gain of distance similarity. Finally, the pattern similarity weights between the corresponding roads at all moments are counted to obtain the sequence of weights for all moments ${{\omega }_{p}}(i,j)$, the adjacency matrix expressed as Eq. \eqref{eq:5} and the matrix visualization is shown in Fig. \ref{fig_3}. a.

\begin{equation}
\label{eq:4}
{{\omega }_{p,t}}(i,j)={{e}^{-\alpha \times dis{{t}_{p,t}}(i,j)}}
\end{equation}

\begin{equation}
\label{eq:5}
{{W}_{p}}=\left[ \begin{matrix}
   0 & {{\omega }_{p}}(1,2) & \cdots  & {{\omega }_{p}}(1,N)  \\
   {{\omega }_{p}}(2,1) & 0 & \cdots  & {{\omega }_{p}}(2,N)  \\
   \vdots  & \vdots  & \ddots  & \vdots   \\
   {{\omega }_{p}}(N,1) & {{\omega }_{p}}(N,2) & \cdots  & 0  \\
\end{matrix} \right]
\end{equation}

\subsubsection{Road Attribute Similarity Graph Network}
The physical spatial and functional distribution of the road networks are discussed above. The statistical distribution of road attributes, i.e., road length, maximum historical flows, and speeds, obtains spatial correlation information. For example, roads with the same maximum flow exhibit similarity in the upper limits of traffic flow. Evaluating the similarity of statistical characteristics in various road attributes sufficiently taps into the spatial correlation, so the influence of the similarity among road attribute distributions should not be ignored. In this paper, the road attribute similarity graph is represented as an undirected graph ${{G}_{s}}=(V,E,{{W}_{s}})$, where the weights ${{\omega }_{s}}(i,j)$ of edges ${{e}_{ij}}$ are the similarity between the road ${{v}_{i}}$ and the road ${{v}_{j}}$ in terms of road attributes, which is calculated as follows:

First, the different dimensions of the full graph traffic data are mapped to the range of [0,1] by the maximum-minimum normalization. Secondly, the attribute sequences of the full graph roads at all moments are developed, and for the roads ${{v}_{i}}$ at moments $t$ are denoted as $av_{t}^{i}=[\max (flow(v_{t}^{i})),\max (speed(v_{t}^{i})),len({{v}_{i}})]\in {{\mathbb{R}}^{3}}$, where $\max (flow(v_{t}^{i}))$, $\max (speed(v_{t}^{i}))$ denote the highest historical traffic flow and the highest average speed, respectively, and $len({{v}_{i}})$ denote the road length. Then, the attribute correlation weights $dis{{t}_{a,t}}(i,j)$ between roads are calculated based on the road attribute sequences by Eq. \eqref{eq:6}, denoted as ${{\omega }_{a,t}}(i,j)$, where $\beta$ denotes the gain used to control the overall attribute similarity. Finally, the weights between roads under all moments are aggregated to obtain the weight sequence ${{\omega }_{a}}(i,j)$, then the adjacency matrix ${{W}_{a}}$ of ${{G}_{a}}$ is expressed as Eq. \eqref{eq:7} and the matrix is visualized as shown in Fig. \ref{fig_3}. b.

\begin{equation}
\label{eq:6}
{{\omega }_{a,t}}(i,j)={{e}^{-\beta \times dis{{t}_{a,t}}(i,j)}}
\end{equation}

\begin{equation}
\label{eq:7}
{{W}_{a}}=\left[ \begin{matrix}
   0 & {{\omega }_{a}}(1,2) & \cdots  & {{\omega }_{a}}(1,N)  \\
   {{\omega }_{a}}(2,1) & 0 & \cdots  & {{\omega }_{a}}(2,N)  \\
   \vdots  & \vdots  & \ddots  & \vdots   \\
   {{\omega }_{a}}(N,1) & {{\omega }_{a}}(N,2) & \cdots  & 0  \\
\end{matrix} \right]
\end{equation}

\subsection{Definition 3: Traffic Condition Grades}
To evaluate the traffic condition objectively and comprehensively, the traffic grades are generated as the coupled representation of various traffic characteristics. Specifically, the Self-Organizing Mapping (SOM) neural network is adopted to couple traffic characteristics, i.e. road average speed, volume and occupancy, as the traffic grades \cite{SOM} \cite{2021An}.

\begin{figure}[!t]
\centering
\includegraphics[width=3in]{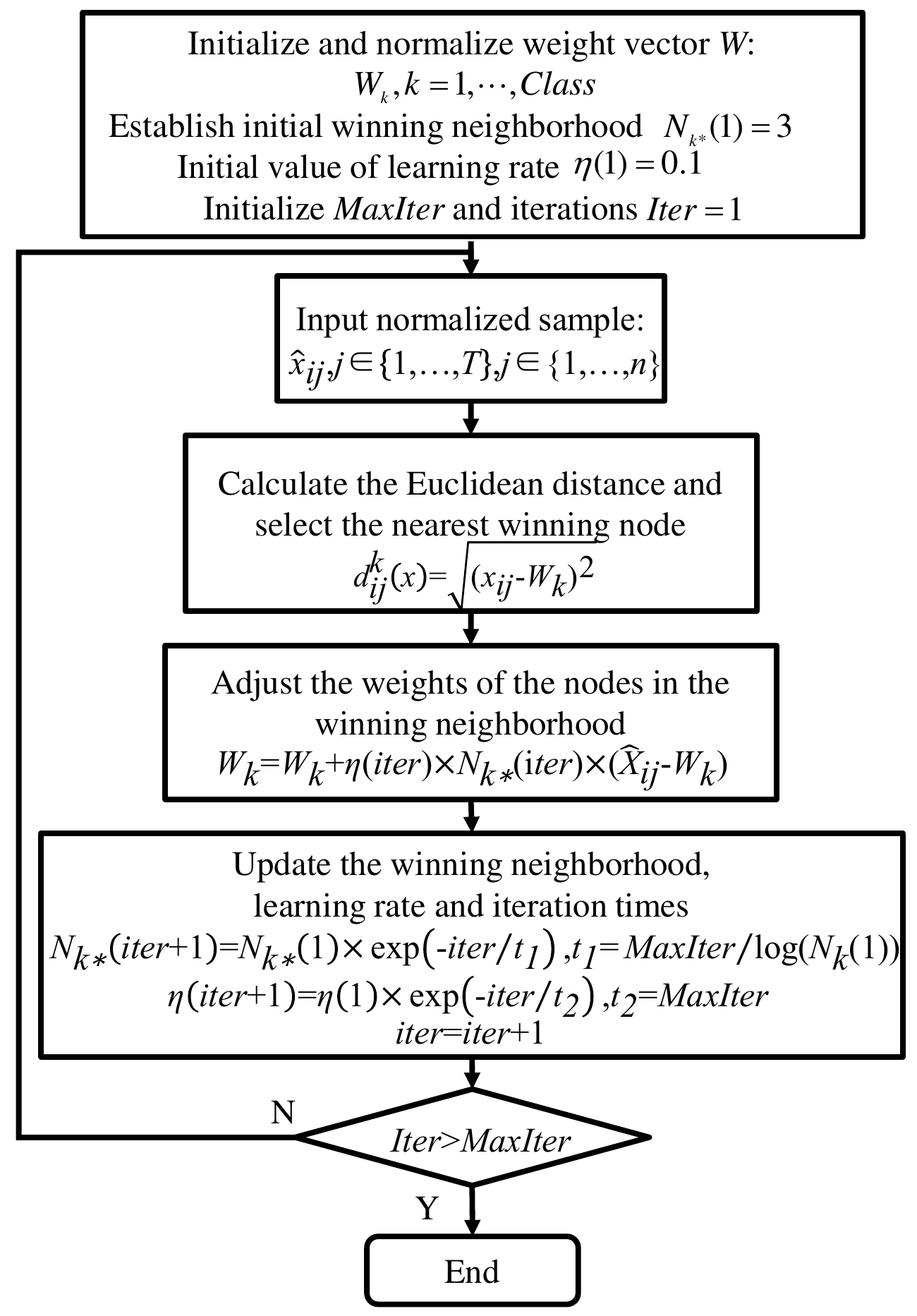}
\caption{Flow chart of SOM network's training process.}
\label{fig_4}
\end{figure}

\begin{figure}[!t]
\centering
\includegraphics[width=1.8in]{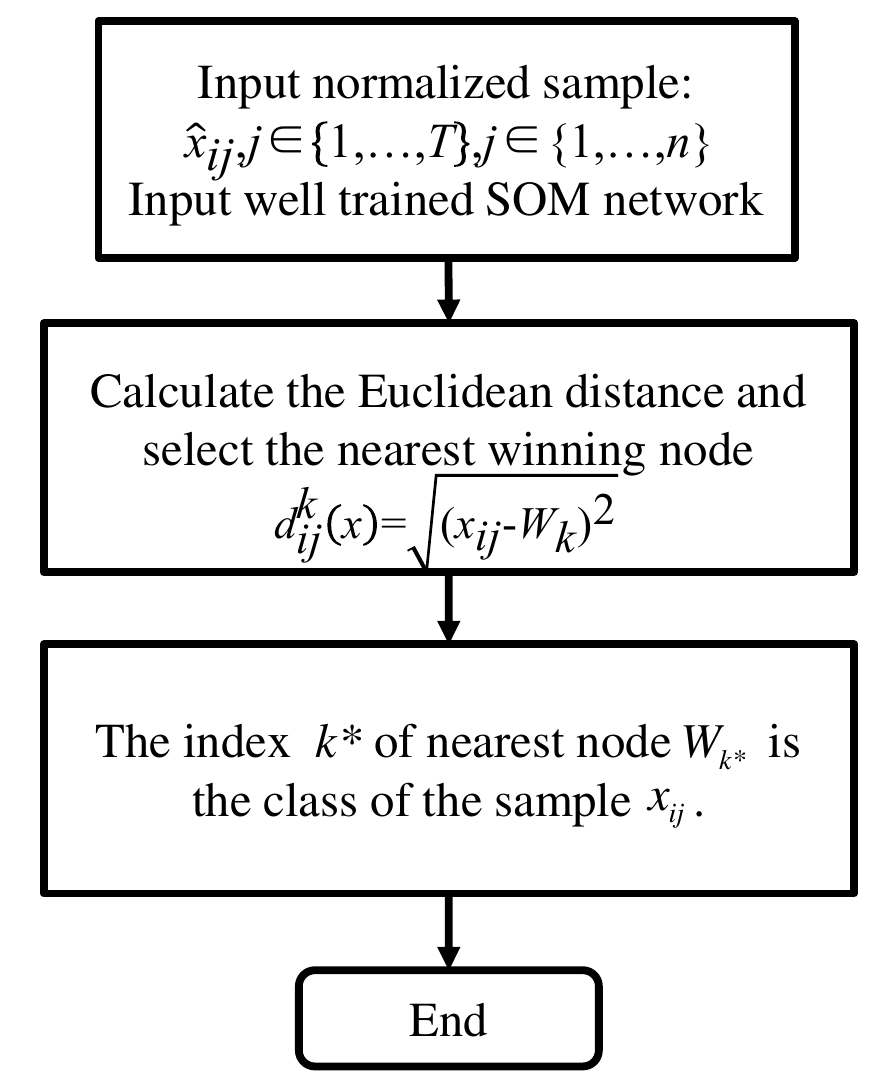}
\caption{Flow chart of SOM network's testing process.}
\label{fig_5}
\end{figure}

The computation process is as follows: traffic state samples under all timestamps in the data are input into the SOM for training, and then the traffic grades of all roads $Y\in [1,\cdots ,Class]$ at total timestamps are obtained based on the trained model for testing, where $Class$ denotes the number of traffic classes. The training and testing process of SOM network is shown as Fig. \ref{fig_4}-\ref{fig_5}.

\subsection{Problem Statement}
\begin{figure}[!t]
\centering
\includegraphics[width=3.5in]{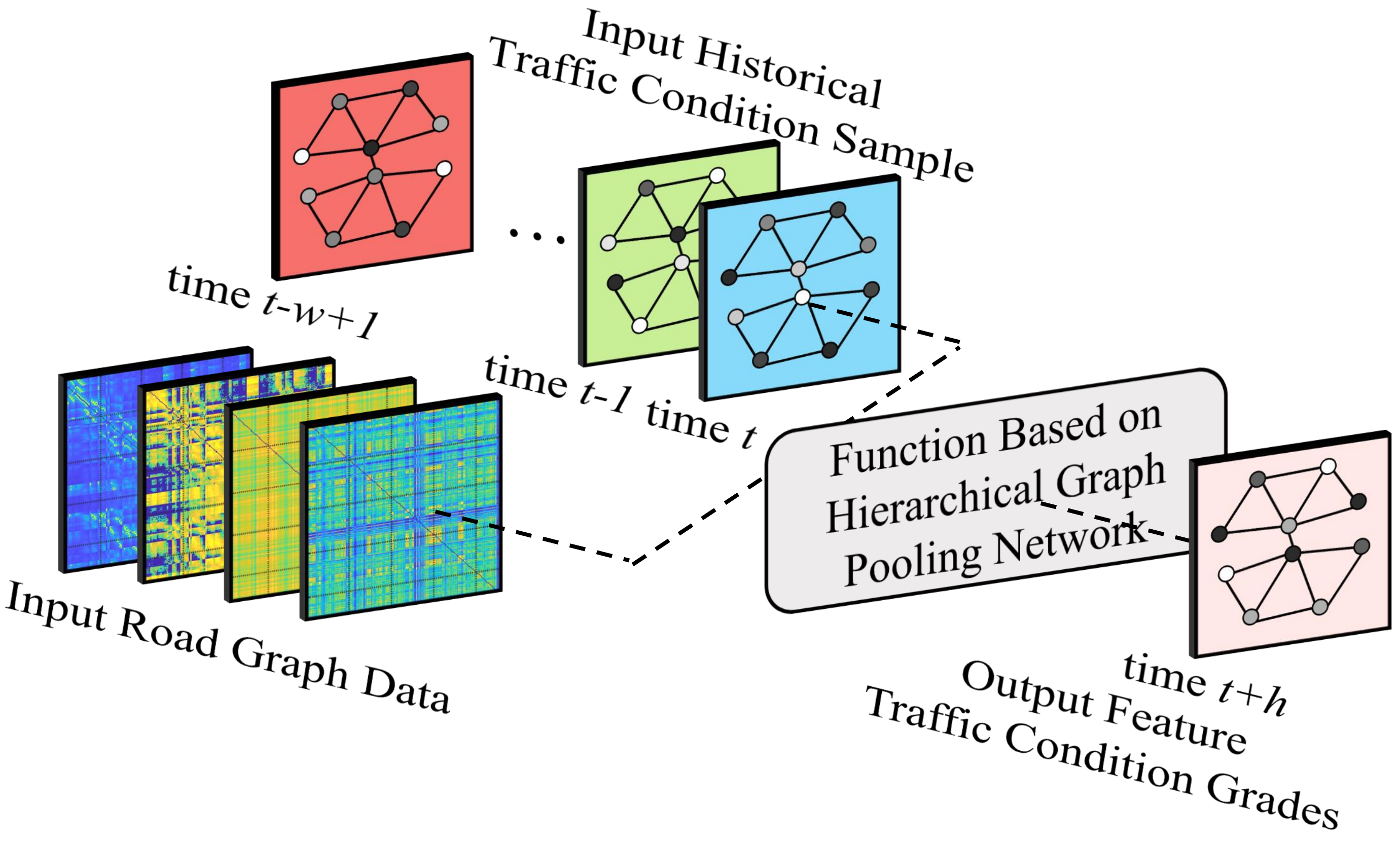}
\caption{The objective of road network-based urban traffic grade prediction.}
\label{fig_6}
\end{figure}

The objective of road network-based urban traffic grade prediction is to learn a prediction function based on a hierarchical graph pooling network, which is shown in Fig. \ref{fig_6}. Specifically, the function can map the historical length $w$ of traffic condition observations ${{X}_{t}}\in {{\mathbb{R}}^{w\times n\times m}}$ at the current moment $t$ to the traffic grade at the future moment $t+{{t}_{p}}$, given different graph network input $G$. Furthermore, $m$ denotes dimensions corresponding to the road's average speed, flow and occupancy, respectively. The function is shown in Eq. \ref{eq:8}.

\begin{equation}
\label{eq:8}
Cl{{s}_{t+h}}={{Y}_{t}}=f(G;{{X}_{t}})
\end{equation}

\section{Method}
Hierarchical graph pooling methods, which have characteristics of multi-level graph representation extraction and redundant information reduction, alleviate the impact of single graph structure and redundant information in GNNs on traffic prediction. To comprehensively describes the process of hierarchical pooling, a general framework is proposed. The hierarchical graph pooling module can be divided into two layers: node feature extraction layer and graph coarsening layer, as shown in Eq. \eqref{eq:9}-\eqref{eq:10}. Firstly, the spatial correlation of the road network based on the corresponding graph is extracted under the node feature extraction layer. Then, the graph coarsening layer coarsens the origin graph network as a smaller one. The spatial correlation under various graph sizes is obtained based on stacking the hierarchical graph pooling modules.

\begin{equation}
\label{eq:9}
{{H}^{(k)}}=Extract({{W}^{(k)}},{{H}^{(k-1)}};{{\theta }^{(k)}}),
\end{equation}

\begin{equation}
\label{eq:10}
W_{{}}^{(k+1)},H{{'}^{(k)}}=Coarsen({{W}^{(k)}},{{H}^{(k)}}),
\end{equation}

\noindent where Eq. \eqref{eq:9} represents the node feature extraction process and $Extract(.)$ denotes the node feature extraction formula. For the layer $k$ feature propagation process, the input layer $k-1$ node embedding ${{H}^{(k-1)}}\in {{\mathbb{R}}^{n\times d}}$, the layer $k$ adjacency matrix ${{W}^{(k)}}\in {{\mathbb{R}}^{n\times n}}$ and the trainable parameters ${{\theta }^{(k)}}\in {{\mathbb{R}}^{d\times d'}}$ are obtained as the layer $k$ node embedding ${{H}^{(k)}}\in {{\mathbb{R}}^{n\times d}}$ after feature extraction. In the initial feature extraction iteration ($k=1$), the input node embeddings ${{H}^{(0)}}$ are initialized by the node features, ${{H}^{(0)}}=F$. $d$ and $d'$ are the node features before and after the extraction of the graph network, respectively.

where Eq. \eqref{eq:10} represents the graph coarsening process and the graph coarsening formula is represented as $Coarsen(.)$. For the graph coarsening process at the layer $k$, the input adjacency matrix ${{W}^{(k)}}\in {{\mathbb{R}}^{n\times n}}$ at the layer $k$ and the feature-extracted node embedding ${{H}^{(k)}}\in {{\mathbb{R}}^{n\times d'}}$ are input, and the coarsened adjacency matrix ${{W}^{(k+1)}}\in {{\mathbb{R}}^{m\times m}}$ and the coarsened node embedding $H{{'}^{(k)}}\in {{\mathbb{R}}^{m\times d'}}$ at the layer $k$ are obtained. $n$ and $m$ represent the number of nodes in the graph network before and after the graph coarsening, respectively, where $m<n$.

Hierarchical pooling methods are divided into node clustering pooling and node drop pooling, depending on the way of graph coarsening. The main idea of node clustering pooling is to consider the graph pool as a node clustering problem and map the nodes into a set of clusters. Then, the clusters are considered as new nodes of a coarsened graph. The classical method for node clustering pooling is Diff Pooling, proposed in NIPS'2018 \cite{2018Hierarchical}. Besides, the designed concept of node drop pooling is to pool graph networks by dropping the nodes with low significance scores and reorganizing the remaining nodes into a new graph. To be specific, the significance scores of nodes are generated by a learnable scoring function. The classical method of node drop pooling is SAG Pooling proposed in ICML'2019 \cite{2019Self}. To sum up, two representative methods of hierarchical graph pooling, namely Diff Pooling and SAG Pooling, are adopted to analyze the performance of the traffic condition prediction task.

\subsection{Differentiable Pooling}
\begin{figure}[!t]
\centering
\includegraphics[width=3in]{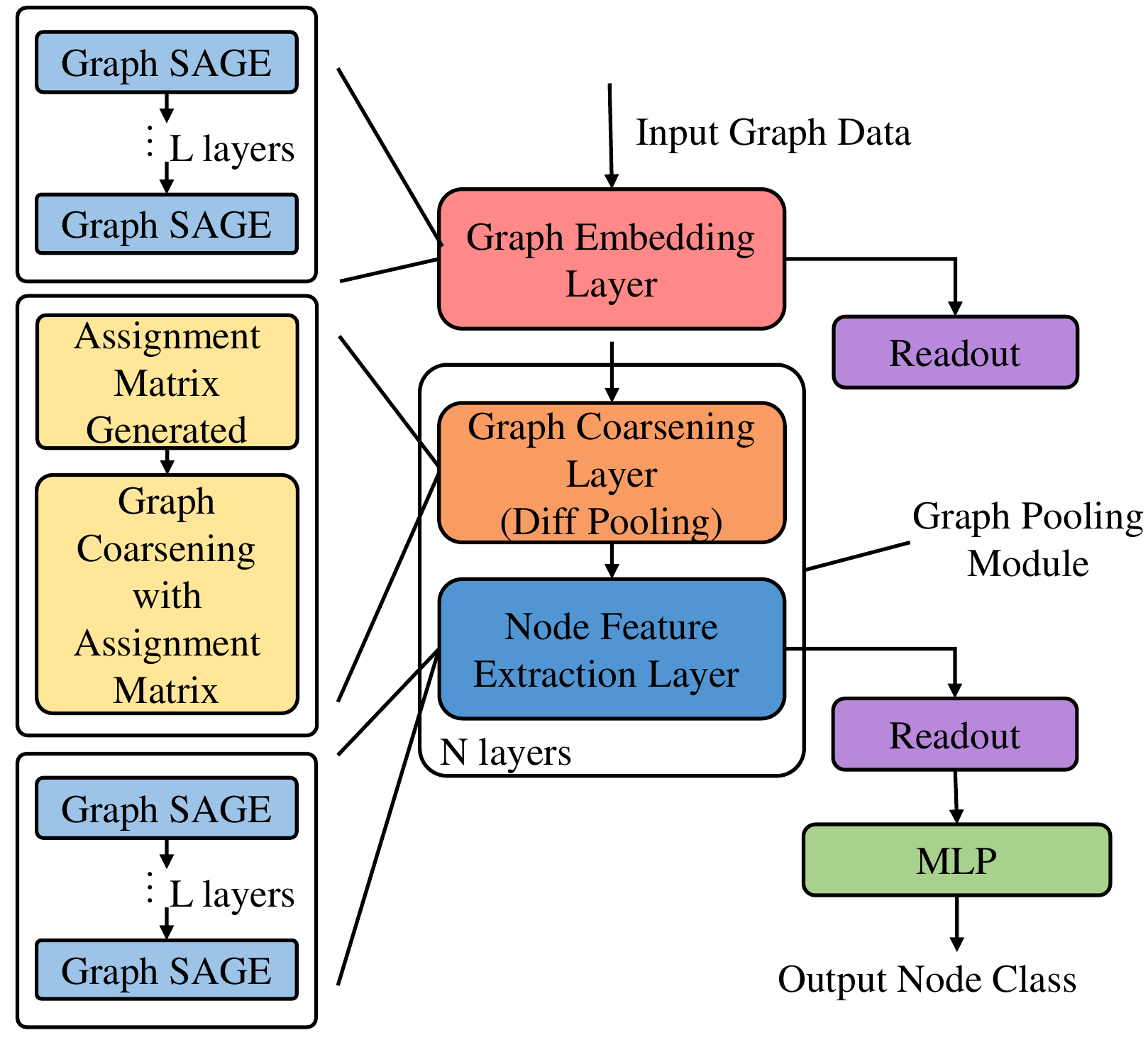}
\caption{The architecture of Diff Pooling method.}
\label{fig_7}
\end{figure}

The overall architecture of Differentiable Pooling (Diff Pooling) obtains the graph embedding layer and stacked $N$ layers graph pooling modules, shown in Fig. \ref{fig_7}. The target of the graph embedding layer is to generate the node embeddings by stacked $L$ layers of Graph SAGE. Besides, the graph pooling module achieves the graph coarsening and graph representation extraction by stacking a graph coarsening layer and a node feature extraction layer. Significantly, the graph coarsening layer is constructed by the operation that generates the assignment matrix and coarsens the graph network based on it. And the node feature extraction layer stacks $L$ layers’ Graph SAGE to extract the spatial correlation from traffic data. The output from the final node feature extraction layer is delivered to the Readout layer to obtain the graph-level representation at the corresponding graph network. Since this paper aims to get the grades of all nodes, the graph-level representations are eventually mapped to all nodes through a multi-layer perceptron.

\begin{figure}[!t]
\centering
\includegraphics[width=2.5in]{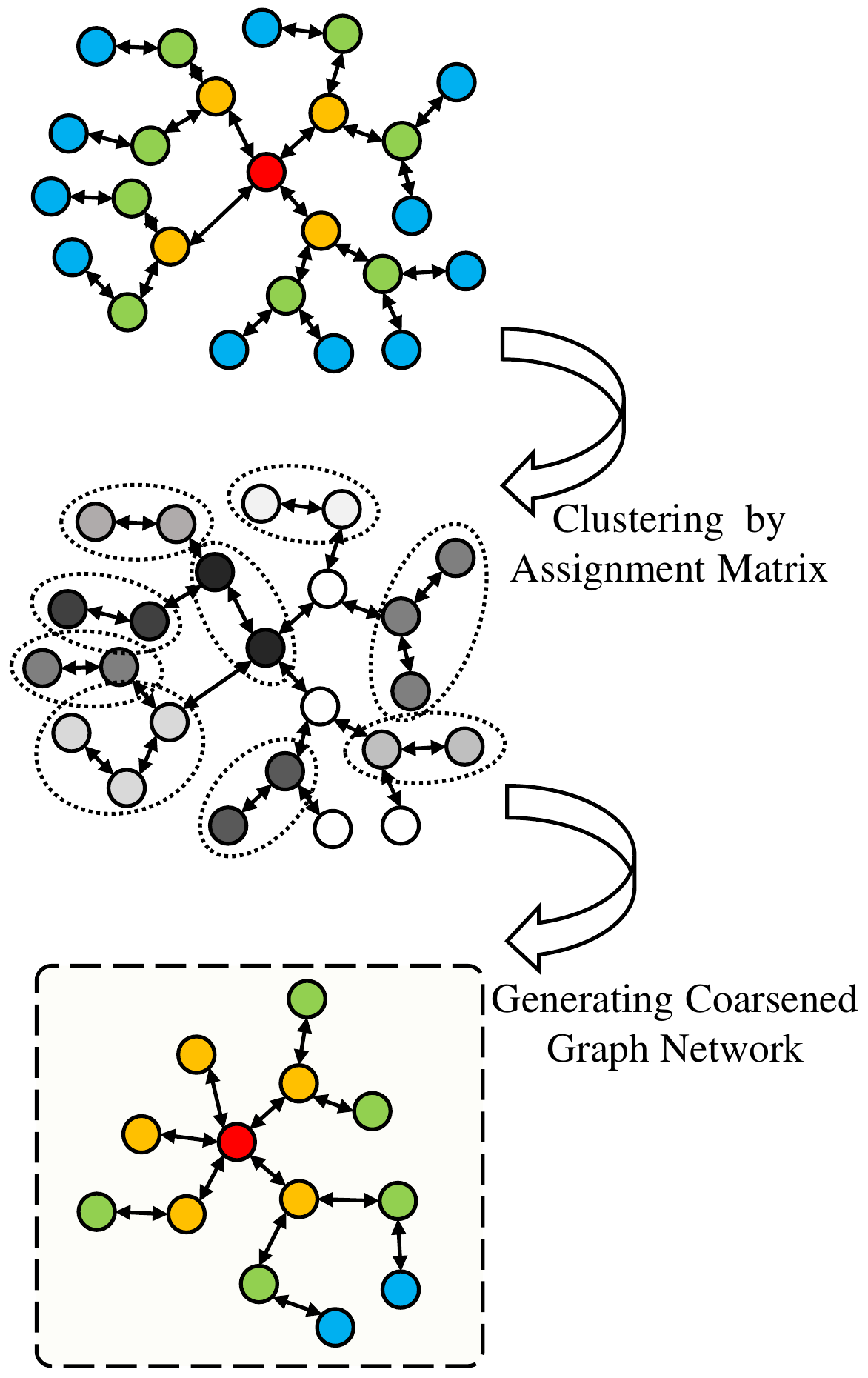}
\caption{The graph coarsening process of the Diff Pooling method.}
\label{fig_8}
\end{figure}

\subsubsection{Graph Coarsening Layer}
In context, the Diff Pooling module coarsens the graph networks by adopting the assignment matrix to cluster nodes, as shown in Fig \ref{fig_8}. Specifically, the generation of the assignment matrix and graph coarsening with the assignment matrix are the core of the graph coarsening layer, which are described in detail below.

Assignment Matrix Generated: In this paper, the process of generating the $l$th layer assignment matrix is introduced: the corresponding assignment matrix ${{S}^{(l)}}$ is obtained by inputting the $l$th layer node feature matrix ${{X}^{(l)}}$ and the adjacency matrix ${{W}^{(l)}}$ into the Graph SAGE network. The assignment matrix represents the assignment weights between the input graph nodes and the clusters of the clusters, and for the $i$th row vector $S_{i}^{(l)}$ represents the correlation between the nodes and each cluster. Then, the allocation matrix is row-regularized, i.e., the operation of $soft\max (.)$, so as to achieve the probabilistic assignment of nodes to clusters, as follows.

\begin{equation}
\label{eq:11}
{{S}^{(l)}}=soft\max (\text{Gra}phSAG{{E}_{l,pool}}({{W}^{(l)}},{{X}^{(l)}}))
\end{equation}

\noindent where $\text{Gra}phSAG{{E}_{l,pool}}$ denotes the Graph SAGE network and ${{S}^{(l)}}\in {{\mathbb{R}}^{{{n}_{l}}\times {{n}_{l+1}}}}$ represents the learnable assignment matrix that measures the probability of node-to-cluster assignment. Each row of the matrix ${{S}^{(l)}}$ corresponds to ${{n}_{l}}$ nodes (or clusters) in the $l$th layer, and each column of ${{S}^{(l)}}$ corresponds to ${{n}_{l+1}}$ clusters in the layer $l+1$. Specifically, the assignment matrix ${{S}^{(l)}}$ assigns all nodes in the $l$th layer to each cluster in the $l+1$th layer.

Graph Coarsening with Assignment Matrix: In the process of layer $l$ graph pooling, the node representations need to be embedded and mapped first, and then the graph coarsening operation is performed based on the node embeddings and the generated assignment matrix. For the generation process of the layer $l$ node embeddings, Diff Pooling applies Graph SAGE network to perform feature aggregation and mapping to node features ${{X}^{(l)}}$. The equation is shown as follow.

\begin{equation}
\label{eq:12}
{{Z}^{(l)}}=GraphSAG{{E}_{l,embed}}({{W}^{(l)}},{{X}^{(l)}})
\end{equation}

\noindent where, $GraphSAG{{E}_{l,embed}}(.)$ represents the Graph SAGE network, ${{Z}^{(l)}}\in {{\mathbb{R}}^{{{n}_{l}}\times {{d}_{l}}'}}$ represents the corresponding node feature embedding. ${{X}^{(l)}}$ and ${{W}^{(l)}}$ represent the node feature matrix and adjacency matrix corresponding to the layer $l$. It is worth noting that the node embedding matrix is generated by the same Graph SAGE network as the assignment matrix. However, the parameters of the two networks are independent of each other.

For the graph coarsening process, based on the already computed assignment matrix ${{S}^{(l)}}$ of the $l$th layer, the node embedding representation ${{Z}^{(l)}}\in {{\mathbb{R}}^{{{n}_{l}}\times d}}$ and the adjacency matrix ${{W}^{(l)}}\in {{\mathbb{R}}^{{{n}_{l}}\times {{n}_{l}}}}$ of this layer ,the graph coarsening process is expressed as Eq. \eqref{eq:13}.

\begin{equation}
\label{eq:13}
({{W}^{(l+1)}},{{X}^{(l+1)}})=DiffPool({{W}^{(l)}},{{Z}^{(l)}})
\end{equation}

\noindent where the coarsen node embedding matrix and the adjacency matrix are obtained by the following two equations:

\begin{equation}
\label{eq:14}
{{X}^{(l+1)}}={{S}^{(l)}}{{Z}^{(l)}}\in {{\mathbb{R}}^{{{n}_{l+1}}\times {{d}_{l+1}}}}
\end{equation}

\begin{equation}
\label{eq:15}
{{W}^{(l+1)}}={{S}^{(l)}}{{W}^{(l)}}{{S}^{(l)}}\in {{\mathbb{R}}^{{{n}_{l+1}}\times {{n}_{l+1}}}}
\end{equation}

In Eq. \eqref{eq:14}, the node embeddings ${{Z}^{(l)}}$ are clustered into ${{n}_{l+1}}$ clusters using the assignment matrix ${{S}^{(l)}}$ to obtain the feature embeddings ${{X}^{(l+1)}}$ of ${{n}_{l+1}}$clusters. Equation \eqref{eq:15} also uses the assignment matrix ${{S}^{(l)}}$ to linearly map the $l$-level adjacency matrix ${{W}^{(l)}}$ to obtain the adjacency matrix ${{W}^{(l+1)}}$ that characterizes the correlation between each cluster.

After graph coarsening, the feature embedding and adjacency matrices of the ${{n}_{l+1}}$ clusters are obtained, where ${{n}_{l+1}}<{{n}_{l}}$. The coarsened graph generated by using node clusters to represent a set of nodes not only reduces the information redundancy in the graph data, but also reduces the computational complexity of the next layer of the graph neural network. It is worth noting that the coarsened adjacency matrix is a real matrix, i.e., a fully connected weighted graph. Each entity in the adjacency matrix $W_{ij}^{(l+1)},i\in {{n}_{l+1}},j\in {{n}_{l+1}}$ represents the correlation weight of the cluster with the cluster. Besides, for the feature embedding matrix of the cluster, each row $X_{i,:}^{(l+1)},i\in {{n}_{l+1}}$ represents the feature embedding of the cluster $i$.

Finally, the coarsened cluster embedding and adjacency matrix are transmitted to the next layer of the graph neural network layer to extract spatial features of relatively high-level graph networks.

\subsubsection{Node Feature Extraction Layer}
The Diff Pooling architecture constructs a node feature extraction layer by stacking layers of the Graph SAGE network to further extract spatial correlation from the coarsened graph network. Compared with the traditional GCN network, Graph SAGE can handle graph data of different sizes, thus gaining scalability. It transfers the goal of the model to learn an aggregator instead of learning a representation for each node, enhancing the flexibility and generalization of the model. Besides, thanks to the flexibility, it is possible to train in batches and improve the convergence speed.

The calculation process of Graph SAGE is as follows: random sampling and feature aggregation of neighbors are performed to reduce the computational complexity and generate target node embeddings, as shown in Eq. \eqref{eq:16}. Specifically, the main aggregation methods are average aggregation, average pooling aggregation, and maximum pooling aggregation. This paper adopts the average pooling $AGGRE_{k}^{\text{a}vg\_pool}$ method \cite{sage} for aggregating neighbor nodes, calculated as in Eq. \eqref{eq:17}. Then, the fused node embeddings are transferred to the fully connected layer to complete the extraction of node features. The calculation process is shown as follows.

\begin{equation}
\label{eq:16}
X_{out}^{l}=GraphSAG{{E}^{l}}(X_{in}^{l},{{W}^{l}})
\end{equation}

\begin{equation}
\label{eq:17}
AGGRE_{k}^{avg\_pool}=mean(\sigma ({{W}_{pool}}h_{u}^{k}+b),\forall u\in (v))
\end{equation}

\noindent where, for the $l$th layer Graph SAGE network $GraphSAGE(.)$, $X_{in}^{l}$ and ${{W}^{l}}$ denote the coarsened node feature matrix and the adjacency matrix, respectively, and $X_{out}^{l}$ denotes the node representation after spatial feature extraction. The aggregation process for any node $u$ is shown in Eq. \eqref{eq:17}, $h_{u}^{k},\forall u\in (v)$ denotes the representation of the neighbor nodes of the target node at the $k$th layer, $mean(.)$ represents the averaging, $\sigma (.)$ represents the sigmoid activation function, ${{W}_{pool}}$ and $b$represents the learnable weight matrix and the bias, respectively.

\subsubsection{Mulit-Layer Perceptron}
After compression and feature extraction, the graph-level representation is finally transmitted into the multi-layer perceptron (MLP) layer to complete the mapping for all grades of the original graph nodes. The MLP consists of stacked layers of fully connected layers with ReLU activation functions, and the output of each layer is used as the input of the next MLP layer, which are calculated as follows.

\begin{equation}
\label{eq:18}
\begin{aligned} & X_{mlp}^{out}=MLP({{X}_{pool}}) \\ & \ \ \ \ \ \ ={{\{\operatorname{Re}LU((W_{linear}^{(l)}\times {{X}_{pool}}+b_{linear}^{(l)}))\}}_{\times L}} \\
\end{aligned}
\end{equation}

\begin{equation}
\label{eq:19}
Linear({{X}_{pool}})={{W}_{l}}
\end{equation}

\noindent where ${{X}_{pool}}$ represents the coarsened graph-level feature representation, $MLP(.)$ denotes the multilayer perceptron model, $\operatorname{Re}LU(.)$ denotes the ReLU activation function, $W_{linear}^{(l)}$ denotes the weight of the linear mapping of layer $l$, and $b_{lienar}^{(l)}$ denotes the bias of the $l$th layer linear mapping, for a total stack of $L$ layers, i.e. $l\in \{1,\cdots ,L\}$. The first two linear layers map the node features to dimensionality ${{d}_{l\_Diff}}$, and the last layer maps to dimensionality $[n\times Class]$, $n$ represents the total number of nodes and $Class$ represents the number of levels. Finally, the output of MLP is operated with $reshape(.)$ and $soft\max (.)$ to obtain the probability matrix of all the nodes corresponding to the traffic grades.

\subsection{Self-Attention Graph Pooling}
\begin{figure}[!t]
\centering
\includegraphics[width=3in]{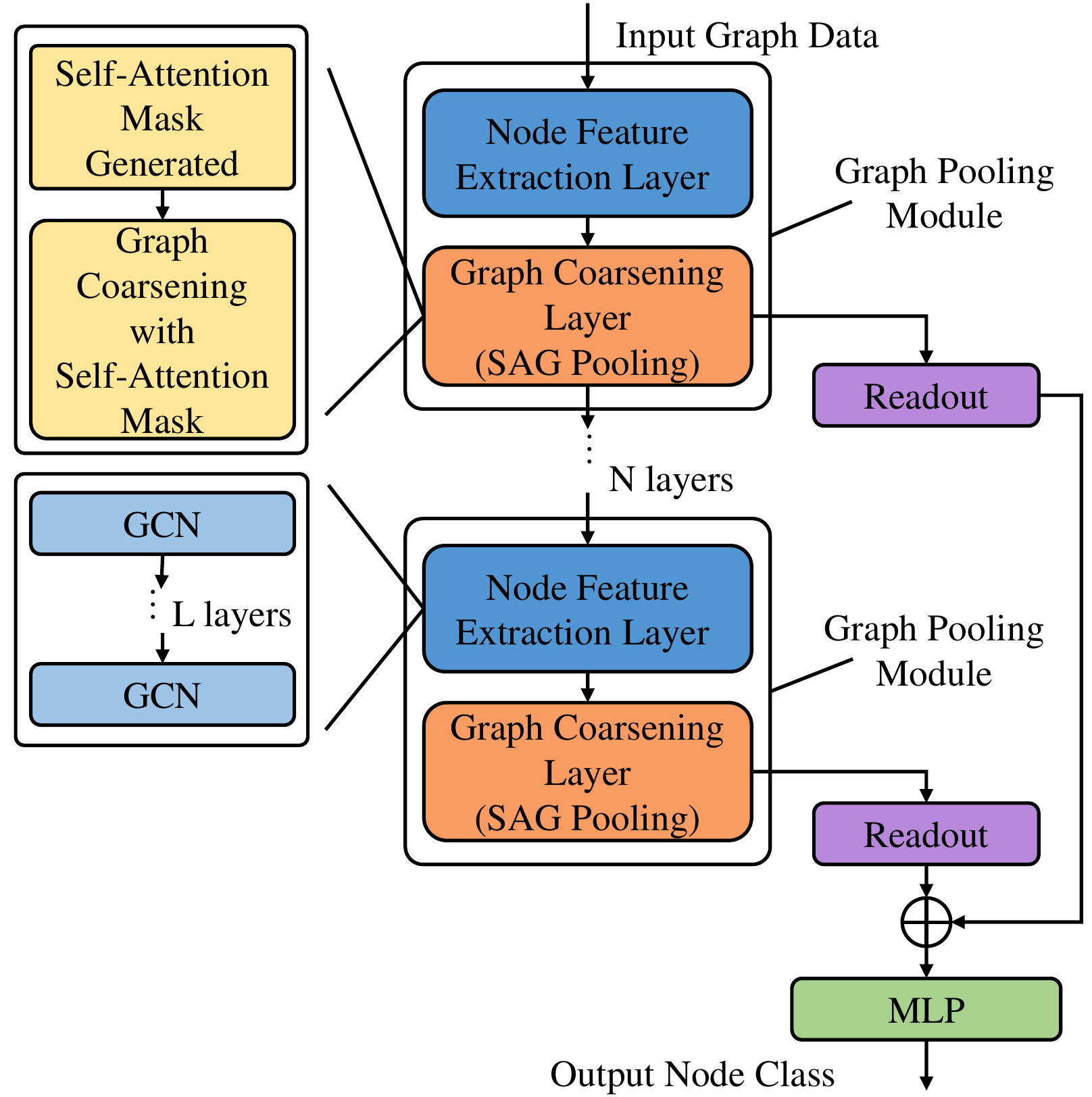}
\caption{The architecture of SAG Pooling method.}
\label{fig_9}
\end{figure}

Self-Attention Graph Pooling (SAG Pooling) is a classical hierarchical pooling method based on node dropping operation. The SAG Pooling architecture is built based on $N$ layers graph pooling module, which is shown in Fig. The graph pooling module consists of node feature extraction layer and graph coarsening layer. particularly, the node extraction layer captures spatial correlation among nodes by stacking $L$ layers GCNs. Besides, the coarsened graph network is generated by SAG Pooling models from graph coarsening layer. The output of each layer of the graph pooling module is transmitted to the next layer on the one hand. On the other hand, the result is delivered to the readout layer for node information aggregation at the corresponding layer. Specifically, the readout layer performs average and maximum pooling for all nodes. Finally, all graph-level representation output from the readout layer is concatenated and transmitted into the MLP layer to obtain the grades at the original graph’s nodes.

\subsection{Self-Attention Graph Pooling}

\subsubsection{Node Feature Extraction Layer}
The SAG Pooling architecture constructs a node feature extraction layer by stacking GCNs, i.e., Graph Convolution Networks (GCNs) proposed by Kipf et al. \cite{2016Semi}. Particularly, GCNs are widely used in traffic prediction tasks \cite{zhao} due to their ability to extract spatial correlations stably and efficiently between nodes in graph networks and their ease of computation. GCNs use a combination of linear transformations and ReLU nonlinear activation functions to achieve aggregation and propagation of node features. The computational formula for the $k$th layer of GCN networks is as follows.

\begin{equation}
\label{eq:20}
\begin{aligned}
  & {{H}^{(k)}}=Extract({{W}^{(k)}},{{H}^{(k-1)}};{{\theta }^{(k)}}) \\ 
 & \ \ \ \ \ \ =\operatorname{Re}LU\left( {{{\tilde{D}}}^{(k)}}{}^{-\frac{1}{2}}{{{\tilde{W}}}^{(k)}}{{{\tilde{D}}}^{(k)}}{}^{-\frac{1}{2}}{{H}^{(k-1)}}{{\theta }^{(k)}} \right) \\ 
\end{aligned}
\end{equation}

\noindent where, ${{\tilde{W}}^{(k)}}={{W}^{(k)}}+I$ represents the adjacency matrix considering the self-loop, ${{\tilde{D}}^{(k)}}=\sum\nolimits_{j}{{{{\tilde{W}}}^{(k)}}_{ij}}$ represents the degree matrix, ${{\theta }^{(k)}}\in {{\mathbb{R}}^{d\times d'}}$ is a trainable parameter matrix, ${{W}^{(k)}}\in {{\mathbb{R}}^{n\times n}}$ represents the adjacency matrix of the graph network, ${{H}^{(k-1)}}\in {{\mathbb{R}}^{n\times d}}$ denotes the node representation of the $k-1$th layer output, and ${{H}^{(k)}}\in {{\mathbb{R}}^{n\times d'}}$ represents the node representation after feature extraction.

\begin{figure}[!t]
\centering
\includegraphics[width=2.5in]{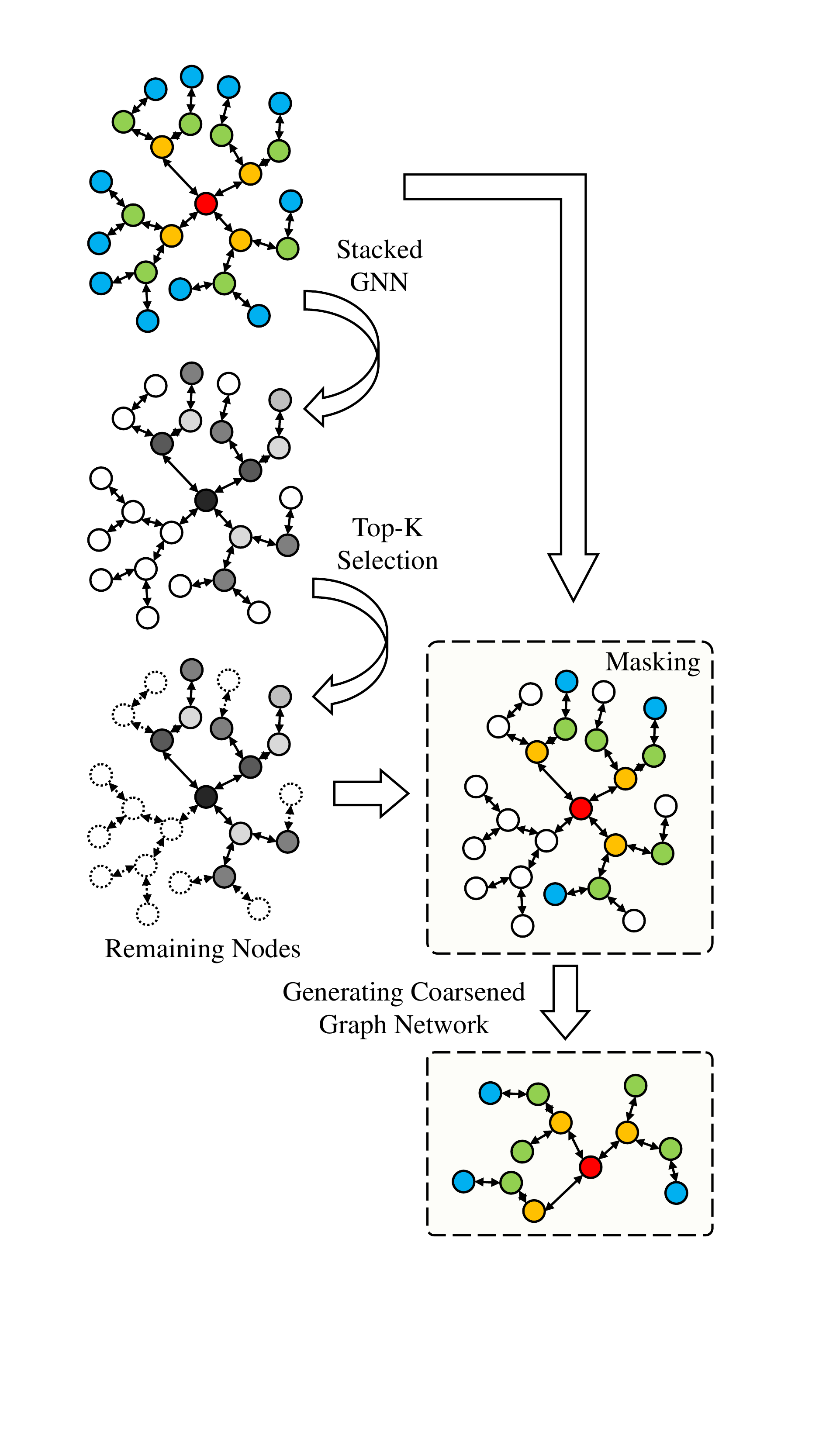}
\caption{The graph coarsening process of the SAG Pooling method.}
\label{fig_10}
\end{figure}

\subsubsection{Graph Coarsening Layer}
The core of the SAG Pooling architecture is the graph coarsening approach in the graph coarsening layer. Specifically, the representation of different node activities is obtained by learning from the graph data, as shown in the Fig. Then the nodes with low activity are masked to complete the dropping of nodes, while the representative nodes are retained and reorganized into a coarsened graph. The critical steps are the generation of self-attentive masks and graph coarsening based on self-attentive masks, which are described in detail below.

Self-Attention Mask Generated: the self-attention mask is generated based on the self-attention mechanism. As a variant of the attention mechanism, the self-attention mechanism can make the network pay more attention to the interaction among the elements within the dataset. The SAG pooling method obtains the correlation among internal elements based on the GCN model, expressed as self-attention scores. The GCN is proposed by kipf \& welling \cite{2016Semi}, as shown below:

\begin{equation}
\label{eq:21}
Z=\sigma \left( {{{\tilde{D}}}^{-\frac{1}{2}}}\tilde{W}{{{\tilde{D}}}^{-\frac{1}{2}}}X{{\theta }_{att}} \right)
\end{equation}

\noindent where, $Z\in {{\mathbb{R}}^{N\times 1}}$ is the self-attention score, $\sigma $ is the activation function, i.e., $\tanh $, $\tilde{W}\in {{\mathbb{R}}^{N\times N}}$ is the adjacency matrix ($\tilde{W}=W+I$) with self-connection, $\tilde{D}\in {{\mathbb{R}}^{N\times N}}$ is the degree matrix of $\tilde{A}$, $X\in {{\mathbb{R}}^{N\times F}}$ is the characteristic matrix of $N$ nodes with $F$ features in the graph network, and ${{\theta }_{att}}$ is the parameter that can be learned in the GCN.

The self-attentive score of each node is obtained by the graph convolution network. Then the nodes are dropped based on the node selection method proposed by Gao et al \cite{2019Graph}. This method ensures that the same size graph is filtered for input graphs of different sizes and structures. The number of nodes is obtained by multiplying the pooling ratio $k\in (0,1]$ with the original number of nodes for the input graph data, i.e., the top $[kN]$ nodes is selected to be retained, as in the Eq. \eqref{eq:22}, thus obtaining a matrix of node screening called the self-attentive mask.

\begin{equation}
\label{eq:22}
idx=top-rank(Z,[kN]),\ {{Z}_{mask}}={{Z}_{idx}}
\end{equation}

\noindent where, $top-rank(.)$ represents the formula that returns the index corresponding to the reserved node, ${{Z}_{idx}}$ represents the attention score of the index, and ${{Z}_{mask}}$ represents the attention mask of the node, which is used for node filtering.

Graph Coarsening with Self-Attention Mask: the node dropping process is as follows: For the coarsened node features, the node feature matrix $X$ is filtered according to the reserved index $idx$, and then the node self-attentive mask is multiplied point by point with the filtered node features . For the coarsened adjacency matrix, the corresponding rows and columns of adjacency matrix are filtered according to the reserved index $idx$. The specific steps are as follows:

\begin{equation}
\label{eq:23}
\left\{ \begin{matrix}
   X'={{X}_{idx,:}}  \\
   {{X}_{out}}=X'\odot {{Z}_{mask}}  \\
\end{matrix} \right.
\end{equation}

\begin{equation}
\label{eq:24}
{{W}_{out}}={{W}_{idx,idx}}
\end{equation}

\noindent where, ${{X}_{idx,:}}$ represents the node characteristic matrix indexed by row, $\odot $ represents Hadamard product, and ${{W}_{idx,idx}}$ represents the adjacency matrix of the corresponding row and column. ${{X}_{out}}$ and ${{W}_{out}}$ respectively represent the coarsened node characteristic matrix and the coarsened adjacency matrix.

\subsubsection{Mulit-Layer Perceptron}
After compression and feature extraction, the graph level-representation is finally transmitted to the multi-layer perceptron (MLP) layer to complete the mapping for all grades of the original graph nodes. The MLP consists of stacked $L$ layers of fully connected layers with ReLU activation functions, and the output of each layer is used as the input of the next layer of MLP. The calculation equations are as follows.

\begin{equation}
\label{eq:25}
\begin{aligned} & X_{mlp}^{out}=MLP({{X}_{pool}}) \\ & \ \ \ \ \ \ ={{\{\operatorname{Re}LU((W_{linear}^{(l)}\times {{X}_{pool}}+b_{linear}^{(l)}))\}}_{\times L}} \\ \end{aligned}
\end{equation}

\begin{equation}
\label{eq:26}
Linear({{X}_{pool}})={{W}_{l}}
\end{equation}

\noindent where ${{X}_{pool}}$ represents the coarsened graph-level representation, $MLP(.)$ represents the MLP model, $\operatorname{Re}LU(.)$ represents the ReLU activation function, $W_{linear}^{(l)}$ and $b_{lienar}^{(l)}$ represent the weight and bias of the $l$th linear mapping layer, respectively. For a total stack of layers, i.e. $l\in \{1,\cdots ,L\}$. The first two linear layers map node features to dimension ${{d}_{l\_SAG}}$, and the last layer maps to dimension $[n\times Class]$, $n$ represents the total number of nodes and $Class$ represents the number of grades. The output of the final MLP is operated with $reshape(.)$ and $Softmax(.)$ to obtain the probability matrix of all nodes corresponding to the traffic grades.

\section{Experiment}
To evaluate classical hierarchical graph pooling methods, we undertake traffic prediction experiments on a real-world traffic dataset. Specifically, the results with the statistical methods and the neural network based methods are compared.

\subsection{Experimental settings}
In the experiments, the data  $PeMS03$, collected by the California Department of Transportation, are used for assessing the methods. $PeMS03$ corresponds to 358 sensors on the California highway system, which measure the average speed, flow, and occupancy of the road per hour. The sensor data from January 1, 2019, to January 31, 2019, is chosen. In addition, a series of graph network inputs are developed in this paper based on the obtained traffic data, which are topological distribution graph (Topo), geographic distribution graph (Geo), historical pattern  graph(HistPatt), and road attribute similarity graph (Attr).

These experiments are performed based on Windows 10 and NVIDIA RTX 3060Ti. The hierarchical graph pooling models and deep-learning baselines are implemented on the Pytorch platform, and the basic machine learning baselines are implemented on scikit-learns platform. The hyperparameters based on the available dataset are formulated in the experiments, as shown in TABLE \ref{tab:table1}-\ref{tab:table2}.

\begin{table}[!t]
\caption{Hyperparameter setting for Diff Pooling method\label{tab:table1}}
\centering
\setlength{\tabcolsep}{1mm}{
\begin{tabular}{c c c c}
\toprule 
\multicolumn{4}{c}{Detailed Parameter Setting}\\
\hline
Number of roads & 358 & Pooling Block layer $N$ & 1 \\
features of roads & 3 & Pooling rate & 0.5 \\
Historical Length & 24 & MLP hidden ${{d}_{l\_Diff}}$ & 320 \\
DTW attenuation rate of Histpatt & 0.1 & Number of road grades & 5 \\
DTW attenuation rate of Attr & 1 & Optimizer & Adam \\
Node embedding & 64 & Learning rate & 5e-4 \\
GCN layer $L$ & 1 & Weight decay & 1e-3 \\
GCN Hidden & 64 & Batch size & 24 \\
Linear Hidden & 1790 & Epoch & 500 \\
\bottomrule 
\end{tabular}}
\end{table}

\begin{table}[!t]
\caption{Hyperparameter setting for SAG Pooling method\label{tab:table2}}
\centering
\setlength{\tabcolsep}{1mm}{
\begin{tabular}{c c c c}
\toprule 
\multicolumn{4}{c}{Detailed Parameter Setting}\\
\hline
Number of roads & 358 & Pooling rate & 0.5 \\
features of roads & 3 & MLP hidden ${{d}_{l\_SAG}}$ & 1790 \\
Historical Length & 24 & Number of road grades & 5 \\
DTW attenuation rate of Histpatt & 0.1 & Optimizer & Adam \\
DTW attenuation rate of Attr & 1 & Learning rate & 5e-4 \\
GCN layer $L$ & 1 & Weight decay & 1e-3 \\
GCN Hidden & 1790 & Batch size & 24 \\
Linear Hidden & 1790 & Epoch & 500 \\
Pooling Block layer $N$ & 3 & & \\
\bottomrule 
\end{tabular}}
\end{table}

\subsection{Evaluations}
In this paper, accuracy (ACC) and quadratic weighted kappa coefficient (KAPPA) are used as the evaluation index of the predictive performance, where the KAPPA indicates the consistency of the predicted grade with the true grade distribution, and this coefficient characterizes the accuracy and deviation of the prediction. The calculation process is based on the confusion matrix, which takes values between -1 and 1. The closer the value is to 1, the higher the consistency of the prediction grade results. The ACC and KAPPA are calculated as follows.

\begin{equation}
\label{eq:27}
ACC=\frac{1}{n}\sum\limits_{t=1}^{n}{(1,\,\ if\,{{v}_{t}}={{{\tilde{v}}}_{t}}\ \text{else}\ 0)}
\end{equation}

\begin{equation}
\label{eq:28}
KAPPA=\frac{{{P}_{o}}-{{P}_{e}}}{1-{{P}_{e}}}
\end{equation}

\begin{equation}
\label{eq:29}
\left\{ \begin{matrix}
   {{P}_{o}}=\sum\nolimits_{i=1}^{Class}{\sum\nolimits_{j=1}^{Class}{{{\omega }_{i,j}}{{p}_{i,j}}}}  \\
   {{P}_{e}}=\sum\nolimits_{i=1}^{Class}{\sum\nolimits_{j=1}^{Class}{{{\omega }_{i,j}}{{p}_{i,:}}{{p}_{:,j}}}}  \\
   {{\omega }_{i,j}}=1-{{\left( \frac{i-j}{Class-1} \right)}^{2}}  \\
\end{matrix} \right.
\end{equation}

\noindent where, for the equation of ACC, i.e., Eq. \eqref{eq:27}, ${{v}_{t}}$ is the true grade, ${{\tilde{v}}_{t}}$ is the predicted rank, and $n$ represents the number of predicted instances. For the equation of KAPPA, i.e., Eq. \eqref{eq:28}-\eqref{eq:29}, $p$ is the confusion matrix, ${{p}_{i,j}}$ is the frequency of predicting roads with true grade $i$ as grade $j$ among all prediction results. ${{p}_{i,:}}$ represents the ratio of the number of instances with true grade $i$ to the total number of instances, ${{p}_{:,j}}$ denotes the ratio of the number of instances with predicted grade $j$ to the total number of instances,  ${{\omega }_{i,j}}$ indicates the weighting weight, and $Class$ represnts the number of ranks.

\subsection{Machine Learning Baselines}
The two graph pooling methods are compared with the following baselines to evaluate the performance of graph pooling methods. Among them, Support Vector Classification (SVC) is basic machine learning method. CNN, GCN, and Graph SAGE are deep learning methods. The average speed and traffic flow of the roads are input into the methods and all baselines are optimized to perform the best performance.

\begin{enumerate}
\item{SVC: The traffic prediction model is trained based on the support vector classification algorithm. Here, we use a radial basis kernel with a regularization parameter of 9.0.}
\item{CNN: a classical CNN for modelling Euclidean spatial data. The network architecture is implemented by stacking three layers of CNN with one layer fully connected model. The three CNN layers have kernel sizes of [10, 5], [6, 3], [4,3], step sizes of [5, 2], [4, 2], [3, 2], number of hidden layer channels of 16, 72, 358, and the final fully connected layer has an output dimension of 5.}
\item{GCN: a widely used model, i.e., GCN proposed by Kipf et al. \cite{2016Semi}, is selected as the representation of Graph Convolution Networks. The model architecture is constructed by stacking three layers of GCN with hidden layer dimension of 64. The output of GCN is transformed into two layers of fully connected model, the first layer maps the feature dimension to 1790, and the hidden dimension of the last layer is 1790, corresponding to 358 nodes with 5 grades.}
\item{Graph SAGE: based on \cite{sage}, similar to the GCN scheme, three layers of Graph SAGE are stacked with specific hidden layer parameters as GCN, and the aggregation method is chosen to average pooling.}
\end{enumerate}

In this experiment, the performance under various predictive lengths and models is compared in the traffic prediction task. Specifically, the different methods are requested to output short-term (1 hour to 3 hours) and long-term (6 hours to 12 hours) traffic grades in the future. The 24 hours of historical traffic conditions are input to the various models. In particular, for GNNs with inputting different types of graph networks, the optimal and average results among them are shown separately in this paper. For graph pooling methods, the best performance among the two mentioned methods is chosen as representation.

\begin{figure}[!t]
\centering
\includegraphics[width=3in]{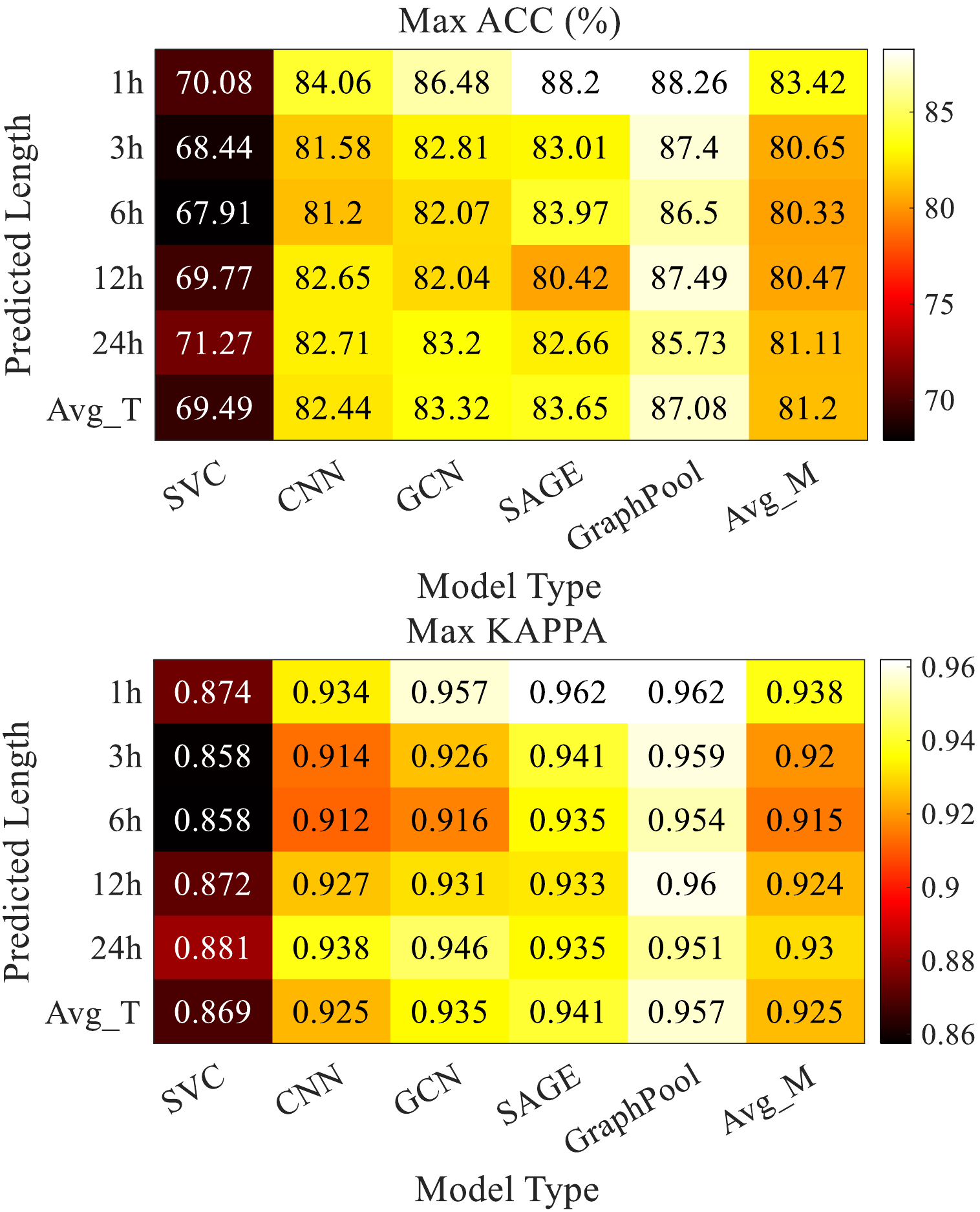}
\caption{Heatmap: baselines Comparison with optimal results based on different graph network inputs.}
\label{fig_11}
\end{figure}

\begin{figure}[!t]
\centering
\includegraphics[width=3in]{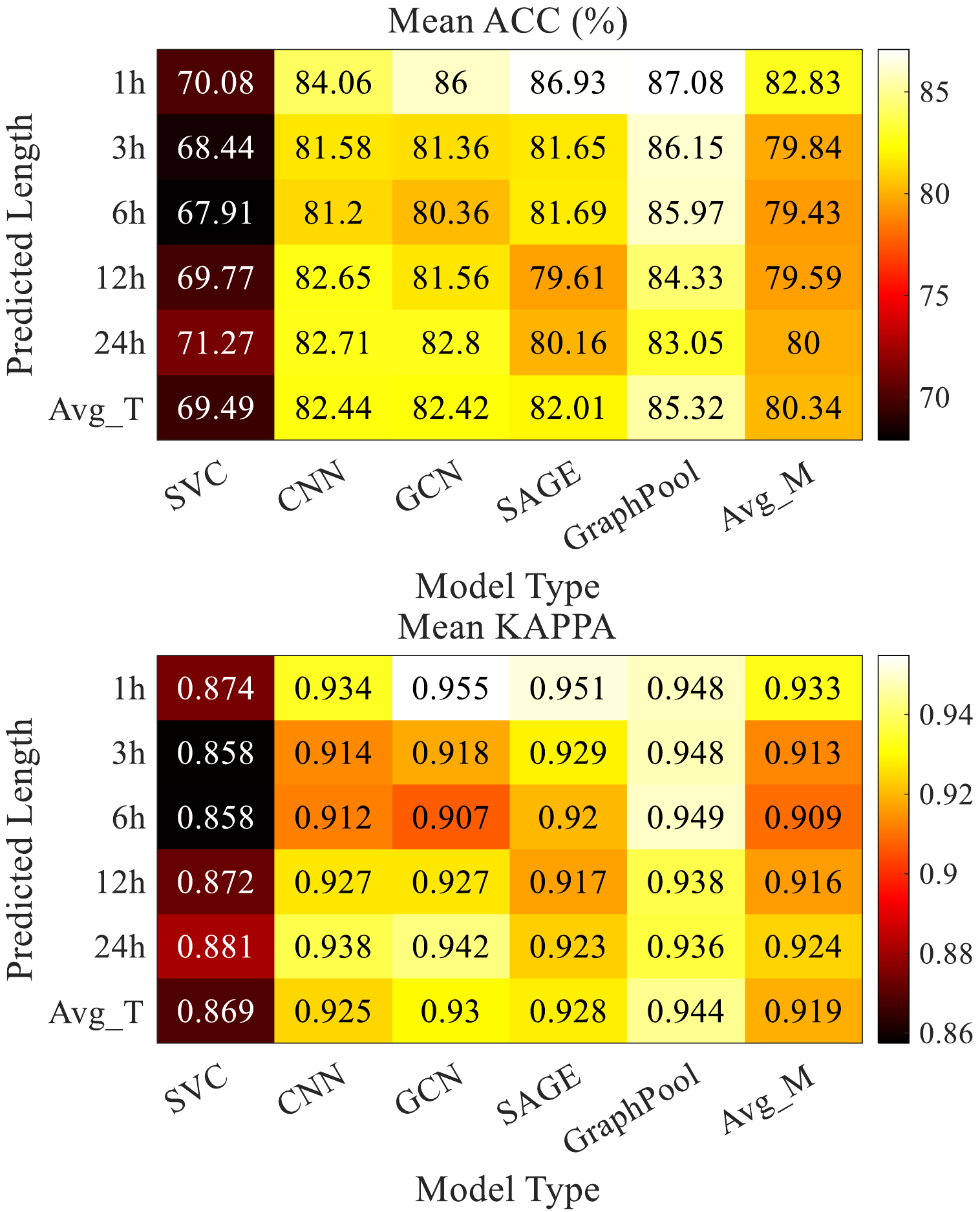}
\caption{Heatmap: baselines Comparison with average results based on different graph network inputs.}
\label{fig_12}
\end{figure}

The prediction effects are shown in the Fig. \ref{fig_11}-\ref{fig_12}, and the methods under different predictive lengths reflect similarity in the prediction performances. Specifically, as the predictive length increases, multiple methods' predictive ACC and KAPPA show a deterioration and back-jump phenomenon. The rightmost column of the best and average effect heatmap offers all methods' average performance at the corresponding prediction lengths. First, the case with a predictive length of 1h shows the highest ACC heatmap. As the predicted length increases to 6h, the predictive accuracy reaches its lowest, decreasing by about 4\% compared to the 1h case. As the predicted length grows to 24h in the future, the ACC picks up and drops by about 3\% compared to the 1h case. Second, in the KAPPA heatmap, as the forecast length grows from 1h to 6h in the future, the KAPPA decreases by about 2.5\%. As the predicted length increases to 24h in the future, KAPPA reduces by about 1\% compared to the 1h in the future case. The daily periodicity of traffic flow can explain this phenomenon. The daily traffic condition distribution has periodicity, which shows that the traffic condition at the forthcoming 24h has a temporal correlation with the current traffic state. Thus, the predicted performance becomes stronger as the predicted length increases to 24h. It is the feature extraction ability of the model with time periodicity that causes this phenomenon to happen.

In addition, the different methods show variability at multiple predictive lengths. As shown in Fig. \ref{fig_11}-\ref{fig_12}, the last row of the four heatmaps indicates the average performance of all methods under different predictive lengths. First, comparing the optimal results based on different graph network inputs in Fig. \ref{fig_11}, SVC performs worst than other deep learning methods at different predicted lengths, with an average ACC and KAPPA of only 69.5\% and 0.869, respectively. CNN improves significantly compared to machine learning methods. The average ACC and KAPPA performance improved by 18.56\% and 6.44\%, respectively, compared to SVC. GCN-like methods, including GCN and Graph SAGE, have improved compared to CNN-like networks, with Graph SAGE improving more significantly. The predictive ACC and KAPPA performance of the Graph Pool method are the best among all methods. Specifically, compared with SAGE, the average ACC and KAPPA are improved by 4.06\% and 1.7\%, respectively, for different predictive lengths, which is a very significant improvement.

Second, average results based on different graph network inputs show similar performance in Fig. \ref{fig_12}. SVC performs as above with CNN-like networks. For the GCN-like methods, the GCN model performs relatively better. Comparing the predictive results of GCN and CNN, it can be found that the performance of ACC and KAPPA is almost the same, which indicates that the impact of different graph network inputs on the GCN-like methods is profound. The Graph Pooling method shows optimal results, specifically, 3.52\% and 1.51\% optimization in accuracy and consistency, respectively. The magnitude of the improvement in the average case is nearly the same compared to the optimal effect case. The phenomenon indicates that Graph Pooling method is less affected by different graph network inputs and have more robust predictive results.

\subsection{Hierarchical Graph Pooling methods}
The above section verifies the effectiveness of hierarchical graph pooling methods by comparing the prediction performance of hierarchical graph pooling methods with the baseline. To further clarify the performance of different hierarchical graph pooling methods under different graph network inputs, two hierarchical graph pooling methods are compared in the following.

\begin{figure}[!t]
\centering
\includegraphics[width=3.5in]{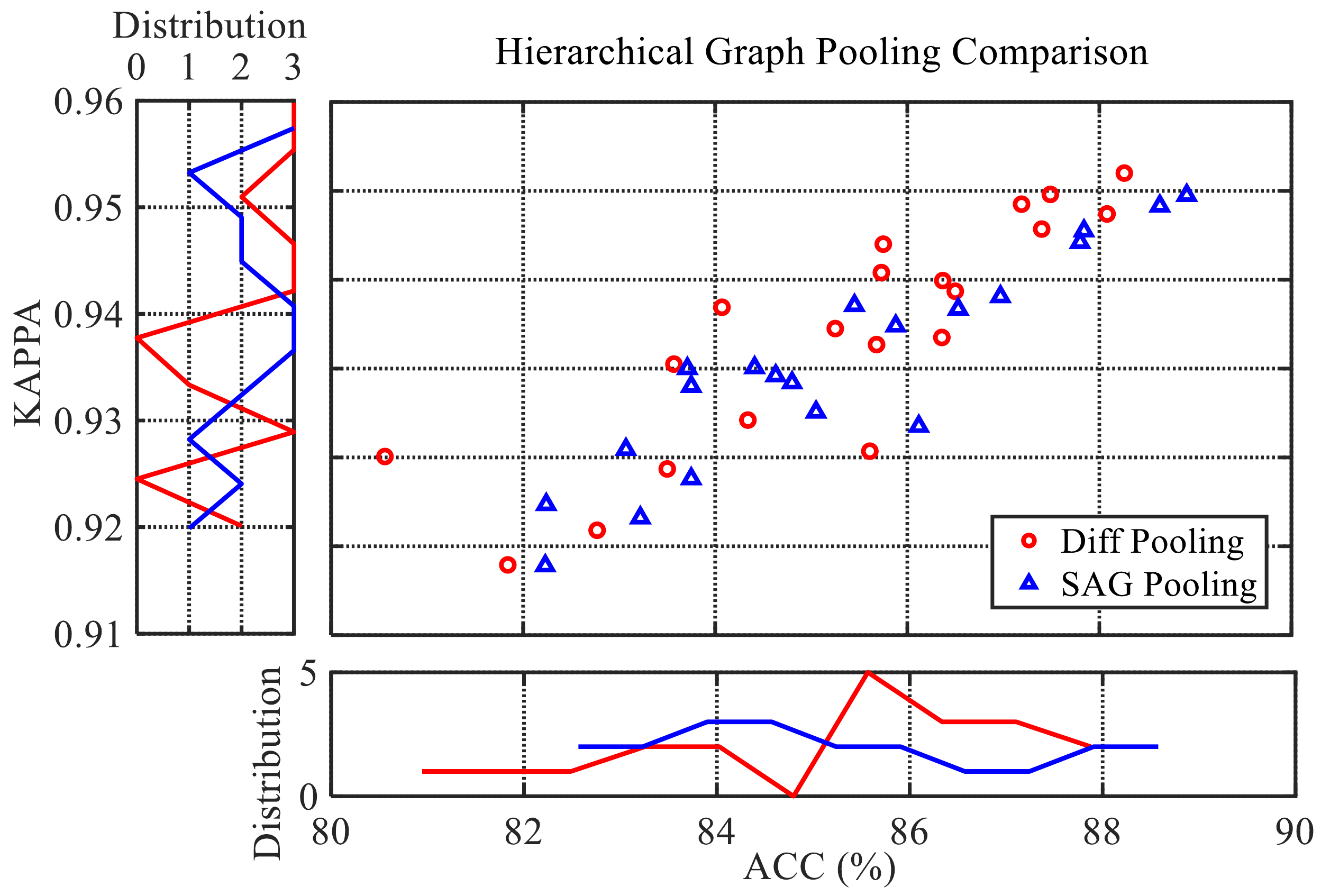}
\caption{Scatter chart: the predictive performance comparison with two hierarchical graph pooling methods.}
\label{fig_13}
\end{figure}

As shown in Fig. \ref{fig_13}, the scatter plots and distributions of ACC-KAPPA for the two hierarchical graph pooling methods based on different graph networks and predictive lengths. First, for accuracy of predicted performance, Diff Pooling has a broader distribution with more distribution in the high accuracy range of 85\% to 88\% and more in the low accuracy range of 81\% to 83\%. SAG Pooling has a more focused accuracy distribution in the medium range, with 83\% to 85\%. Second, for the consistency of predictive performance, Diff Pooling is more distributed in the field of high Kappa with 0.94 to 0.96, and SAG Pooling has more distribution in the range of medium KAPPA with 0.93 to 0.94. Finally, the performance of the two hierarchical graph pooling methods is similar, as shown by the distribution of scatter points. The difference in distributions between the two methods is that the Diff Pooling method is significantly more distributed in the higher accuracy range. Additionally, the SAG Pooling method is distributed more in the lower accuracy and consistency ranges.

\subsection{Structure of Graph Network}
For GNNs, the above focuses on the comparison from a methodological perspective. However, the impact of different graph network inputs on GNNs cannot be ignored, and this section focuses on graph network inputs to show the impact on the prediction effect.

\begin{figure}[!t]
\centering
\includegraphics[width=3in]{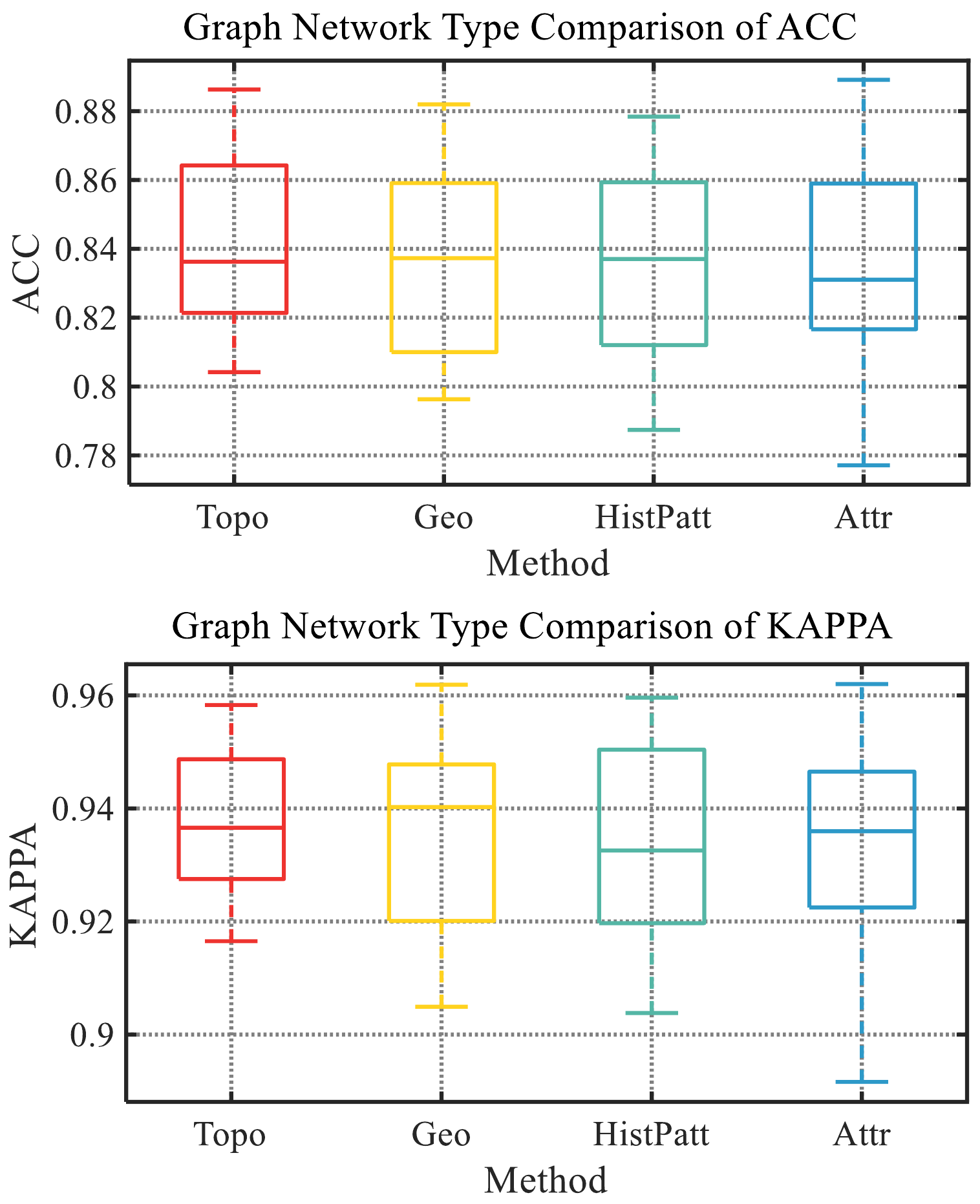}
\caption{Box chart: the predictive performance comparison with different graph network inputs.}
\label{fig_14}
\end{figure}

This paper records the results of two GCN-like methods and two graph pooling methods based on different graph network inputs. The accuracy and consistency of predictive effects are shown in Fig. \ref{fig_14}. For the topological distribution graph, the distributions in both ACC and KAPPA are more concentrated, while the lower limit of the prediction effect is the highest at 0.815. For the geographical distribution graph, the trend of accuracy distribution is slightly worse than that of the topological distribution graph, and the distribution of its KAPPA is relatively scattered from 0.915 to 0.965. For the historical pattern graph, the trend of ACC and KAPPA distribution is similar to that of the geographical distribution graph with a lower limit at 79\%. The median of KAPPA is the weakest among all graph types at 0.93. For the road attribute similarity graph, the distribution of ACC and KAPPA is the most scattered, but the interquartile spacing is small.

\begin{figure}[!t]
\centering
\includegraphics[width=3in]{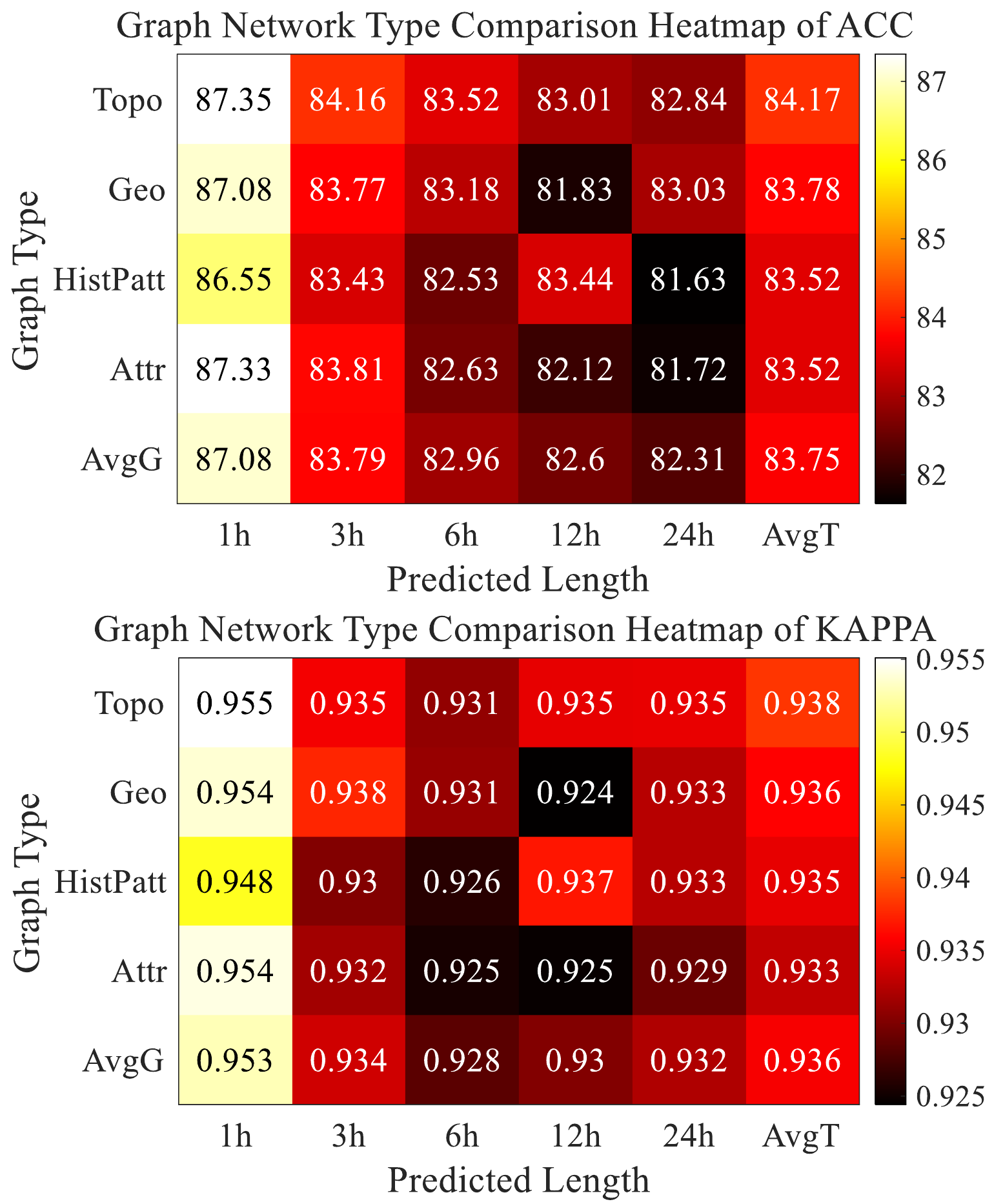}
\caption{Heatmap: the predictive performance comparison with different graph network inputs.}
\label{fig_15}
\end{figure}

By analyzing the statistical distribution based on different graph network inputs, we summarize the overall effect of varying graph networks on the prediction effect. A heatmap of the predicted results of different graph network inputs at different predictive lengths is constructed, which is shown in Fig. \ref{fig_15}. First, the rightmost column in the figure indicates the average performance of all methods at implausible predictive lengths. The topological and geographical distribution graphs have better results than other maps networks, with ACC improvement of about 1\% and KAPPA improvement of about 0.5\%. Secondly, the bottom row of the figure indicates the average of all methods under the corresponding predictive length, which further verifies that the distribution trend of the prediction effect shows a downward convex curve shape as the predictive length increases. Finally, observing the specific impact of different methods under each predictive length, the topological distribution graph and road attribute similarity graph are more suitable for predicting traffic states with lengths of 1h and 3h with about 1\% improvement in ACC and about 0.7\% improvement in KAPPA. For long-term forecasting with a forecast length of 12h, the topological distribution graph shows better results than the historical pattern graph, with ACC improvement of about 1.5\% and KAPPA improvement of about 0.5\%. In the case of predicting the future 24h, the topological distribution graph and the geographic distribution graph reflect the advantages with ACC improvement of about 1.5\% and KAPPA improvement of about 0.5\%.

\section{Discussion}
The above-mentioned experimental sections are compared and tested from three perspectives: baseline, hierarchical pooling method, and network structure. The in-depth information inherent in the results has a guiding role in the task of traffic level prediction, and the following paper analyzes the laws based on these three perspectives in detail:

\subsection{Comparision with Baselines}
By comparing different learning baselines, firstly, the deep learning model improves significantly compared with the machine learning model, indicating that deep learning has strong modelling ability in spatial correlation of traffic data. In contrast, basic machine learning cannot realize the modelling of spatial characteristics due to its simple structure. Among the deep learning models, the GCN-like models for non-Euclidean spatial modelling are more suitable for mining hidden information in traffic data than CNNs for Euclidean spatial correlation modelling, further proving that traffic data exhibit non-Euclidean characteristics in space. Finally, comparing GCN-like models, the graph pooling models are significantly better than traditional GCN models, proving that the spatial correlation in a single network learned by traditional GCN models is limited. Graph coarsening effectively synthesizes a multi-level and multi-scale graph network, thus capturing complex spatially correlated attributes in traffic data more effectively.

\subsection{Comparision with Graph Pooling Methods}
Based on the comparison of the two hierarchical pooling methods, traffic prediction patterns can be found. The prediction performance of Diff Pooling is relatively more distributed in medium-high accuracy with medium-high kappa coefficients, while the distribution of the predicted performance of SAG Pooling is relatively flatter. The predicted performance’s difference reflects the characteristics of different graph pooling methods. Diff Pooling coarsens the graph network by clustering, losing less information. Nevertheless, the reasonable use of the graph network information depends on the clustering results. Different clustering results lead to a significant difference, resulting in relatively unstable results. SAG Pooling coarsens the graph network by node dropping. Although some graph network information is inevitably lost, it can ensure that the nodes with a large amount of information are retained, and the nodes with little information are discarded, making its prediction performance more stable.

\subsection{Comparision with Graph Netowrk Structure}
By summarizing the comparative experimental results of different graph network inputs, the following conclusions can be drawn: first, topological distribution maps perform very robustly in both short-term and long-term forecasting, which indicates that topological distributions can appropriately respond to spatial correlations, and such simple topological relationships do not favour the representation of spatial connections or weaken the temporal relations embedded in spatial distributions. Second, the geographic distribution and the road genus similarity map are more suitable for short-term forecasting tasks. Geographical distribution adjusts topological connections by considering road length, while road attribute similarity divides edges from fixed attributes such as road length. These graph networks reinforce spatial relationships based on rich geographic information. This enhanced spatial correlation is advantageous for short-term forecasting, as shown by the fact that neighboring roads have similar traffic states shortly. However, the long-term forecasting task relies more on the temporal information embedded in the spatial distribution, as shown by the high probability that roads with similar temporal trends will have the same performance in the long-term future. This enhanced spatial information strengthens the importance of spatial information and weakens the weight of temporal information to a certain extent, which leads to the geographic distribution graph and road attribute similarity graph being more suitable for short-term forecasting. Finally, the historical pattern graph defines spatial distribution by historical time-series similarity. This temporal trend does not focus on recent changes but considers periodic changes over a long period, which leads to its poor performance in short-term forecasting tasks that rely on spatial linkages, while it performs better in long-term forecasting tasks that depend on temporal trend changes.

Based on the above laws, the traffic forecasting task can be better guided. However, there are some shortcomings in the current method analysis:

\begin{enumerate}
\item{This paper is only based on the data collected by road network sensors for learning and analysis without considering the weather and event factors in traffic.}
\item{This paper mainly analyzes the predicted effect of different models but lacks the calculation and analysis of the computational efficiency and memory consumption of the models.}
\end{enumerate}

Therefore, considering external factors, such as weather and events, and increasing the analysis of computational efficiency are future directions for improvement. Among them, for extracting useful traffic features from external factors, we consider adding a CNN model-based external factor feature extraction module to extract traffic features.

\section{Conclusion}
This paper applies the mainstream hierarchical graph pooling methods to the traffic prediction task. The graph pooling methods are used to learn different grades of spatial information, and the effectiveness of the methods is verified in comparison with the baseline. Then, the graph pooling methods and graph network construction experiments are conducted to analyze and summarize some laws of the graph pooling method for the traffic prediction task.

Despite the promising results of the graph pooling method, there is still some space for optimization in improving and analyzing the graph pooling method. Specifically, analyzing traffic prediction considering environmental factors and computational efficiency is a hot topic. Incorporating these improvements can provide a more powerful and robust model with more detailed analyses in traffic state prediction to achieve higher performance.

\bibliographystyle{IEEEtran}
\bibliography{ref}

\newpage
\begin{IEEEbiography}[{\includegraphics[width=1in,height=1.25in,clip,keepaspectratio]{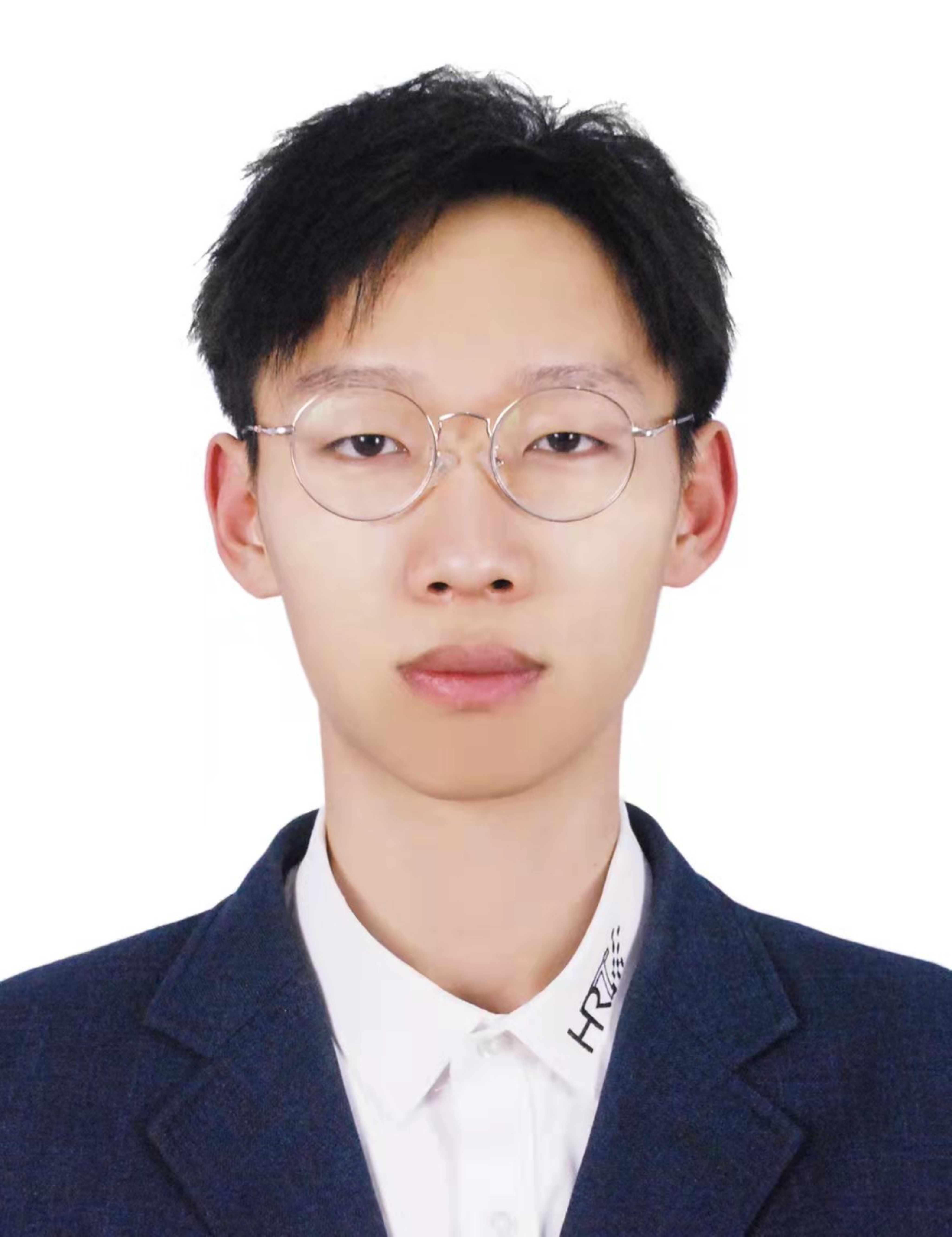}}]{Shilin Pu} was born in Hulunbuir City, Inner Mongolia Autonomous Region on June 25th, 1998. He received the B.E. degree in vehicle engineering from Harbin Institute of Technology, Weihai, China. He is currently pursuing continuous academic program that involves both postgraduate study in mechanic engineering with Jilin University, Changchun, China. 

His research interests include deep learning, Traffic flow forecasting, optimal energy management strategy about plug-in hybrid vehicles.

\end{IEEEbiography}
\vspace{-10mm}
\begin{IEEEbiography}[{\includegraphics[width=1in,height=1.25in,clip,keepaspectratio]{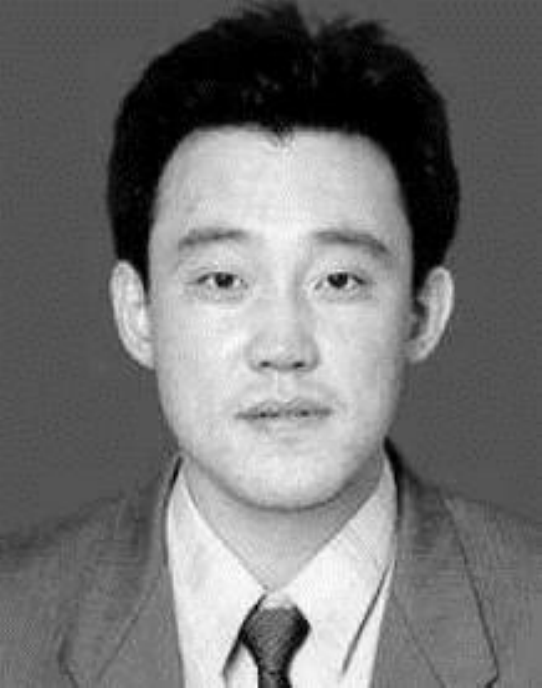}}]{Liang Chu} was born in 1967. He received the B.S., M.S., and Ph.D. degrees in vehicle engineering from Jilin University, Changchun, China. He is currently a Professor and the Doctoral Supervisor with the College of Automotive Engineering, Jilin University. His research interests include the driving and braking theory and control technology for hybrid electric vehicles, which conclude powertrain and brake energy recovery control theory and technology on electric vehicles and hybrid vehicles, theory and technology of hydraulic antilock braking and stability control for passenger cars, and the theory and technology of air brake ABS, and the stability control for commercial vehicle. 

Dr. Chu has been a SAE Member. He was a member at the Teaching Committee of Mechatronics Discipline Committee of China Machinery Industry Education Association in 2006. 
\end{IEEEbiography}
\vspace{-10mm}
\begin{IEEEbiography}[{\includegraphics[width=1in,height=1.25in,clip,keepaspectratio]{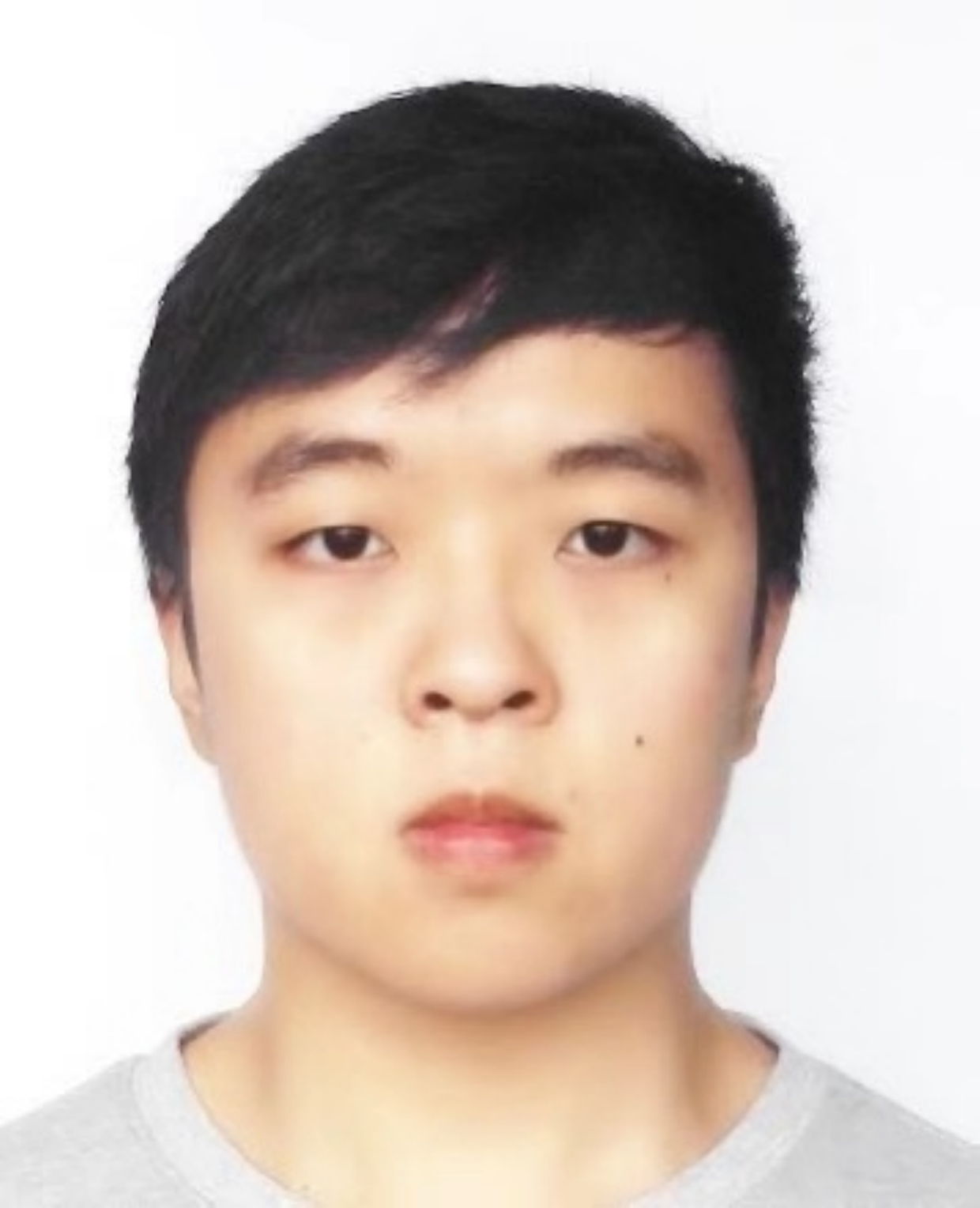}}]{Zhuoran Hou} received the B.S. degree in vehicle engineering from Chongqing University, Chongqing, China, in 2017. and the M.S. in Automotive Engineering from Jilin University, China, in 2020. He is currently pursuing continuous academic program involving doctoral studies in automotive engineering with Jilin University, Changchun, China. 

His research interests include basic machine learning, optimal energy management strategy about plug-in hybrid vehicles. 
\end{IEEEbiography}
\enlargethispage{-9cm}

\newpage
\begin{IEEEbiography}[{\includegraphics[width=1in,height=1.3in,clip,keepaspectratio]{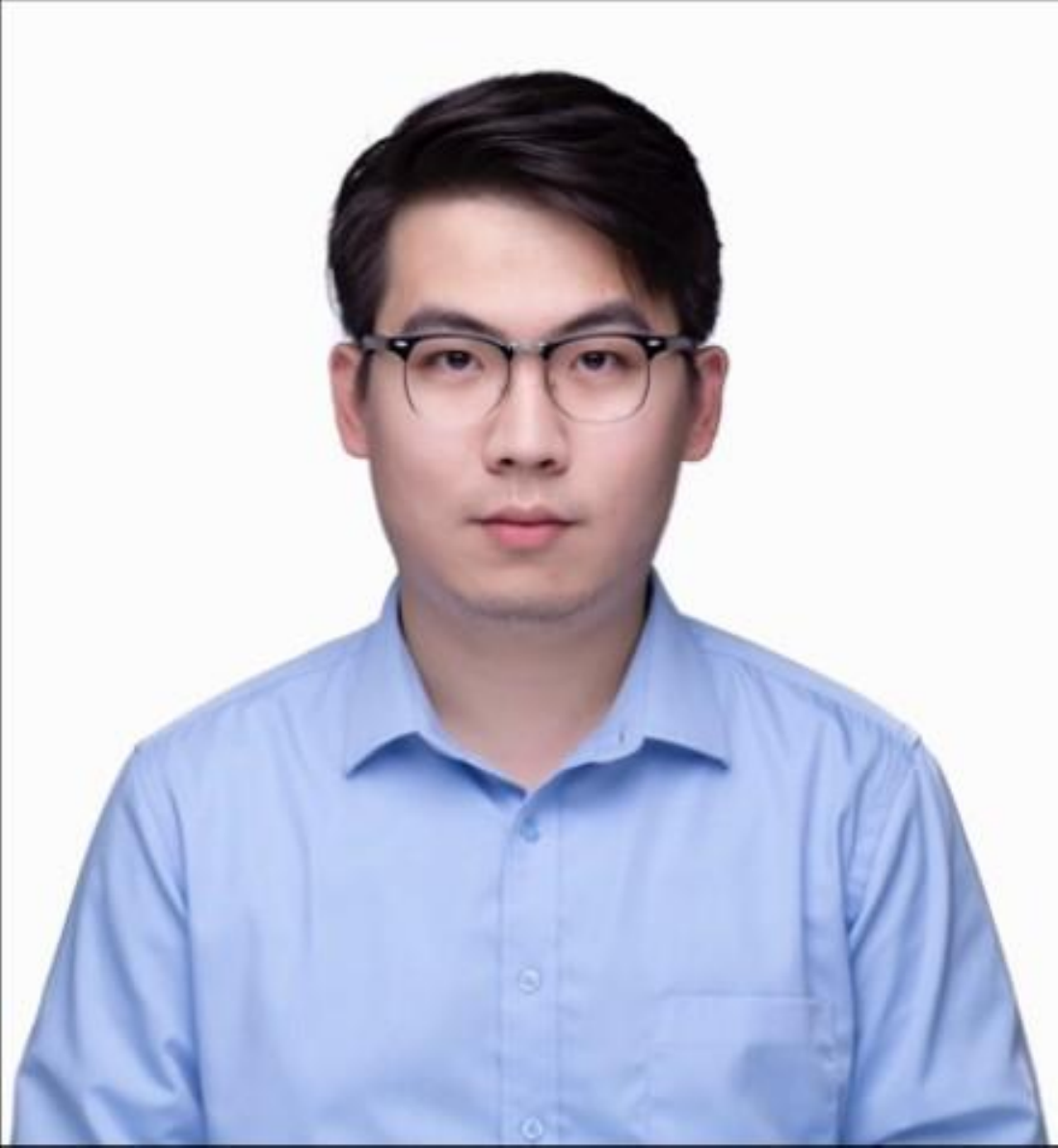}}]{Jincheng Hu} (Student Member, IEEE) received the B.E. degree in information security from the Tianjin University of Technology, Tianjin, China, in 2019 and the M. Sc. degree in information security from the University of Glasgow, Glasgow, UK, in 2022. He is currently working toward the Ph.D. degree in Automotive with the Loughborough university, Loughborough, UK. 

His research interests include reinforcement learning, deep learning, cyber security, and energy management.
\end{IEEEbiography}
\vspace{-10mm}

\begin{IEEEbiography}[{\includegraphics[width=1in,height=1.25in,clip,keepaspectratio]{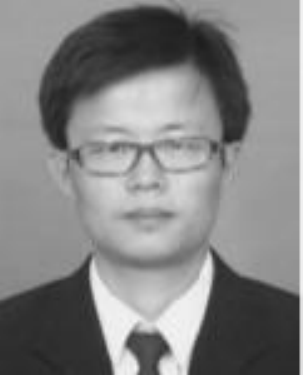}}]{Yanjun Huang} (Member, IEEE) received the Ph.D. degree in mechanical and mechatronics engineering from the University of Waterloo, Waterloo, Canada, in 2016.

He is currently a Professor with the School of Automotive Studies, Tongji University, Shanghai, China. He has authored several books and over 50 papers in journals and conferences. His research interests include the vehicle holistic control in terms of safety, energy saving, and intelligence, including vehicle dynamics and control, hybrid electric vehicle/electric vehicle optimization and control, motion planning and control of connected and autonomous vehicles, and human-machine cooperative driving. Dr. Huang serves as the Associate Editor and Editorial Board Member for the IET Intelligent Transport System, Society of Automotive Engineers (SAE) International Journal of Commercial vehicles, International Journal of Vehicle Information and Communications, Automotive Innovation, etc.
\end{IEEEbiography}
\vspace{-10mm}

\begin{IEEEbiography}[{\includegraphics[width=1in,height=1.25in,clip,keepaspectratio]{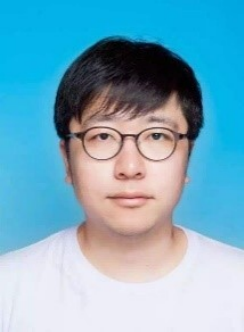}}]{Yuanjian Zhang} (Member, IEEE) received the M.S. in Automotive Engineering from the Coventry University, UK, in 2013, and the Ph.D. in Automotive Engineering from Jilin University, China, in 2018. In 2018, he joined the University of Surrey, Guildford, UK, as a Research Fellow in advanced vehicle control. From 2019 to 2021, he worked in Sir William Wright Technology Centre, Queen’s University Belfast, UK. 

He is currently a Lecturer with the Department of Aeronautical and Automotive Engineering, Loughborough University, Loughborough, U.K. He has authored several books and more than 50 peer-reviewed journal papers and conference proceedings. His current research interests include advanced control on electric vehicle powertrains, vehicle-environment-driver cooperative control, vehicle dynamic control, and intelligent control for driving assist system.
\end{IEEEbiography}
\enlargethispage{-7.7cm}

\end{document}